%% file: main.tex
\documentclass[letterpaper]{article} 
\usepackage{aaai24}  
\usepackage{times}  
\usepackage{helvet}  
\usepackage{courier}  
\usepackage[hyphens]{url}  
\usepackage{graphicx} 
\usepackage{natbib}  
\usepackage{caption} 
\usepackage{algorithm}
\usepackage{algorithmic}
\usepackage{amsmath}
\usepackage{amssymb}
\usepackage{amsthm}
\usepackage{braket}
\usepackage[utf8]{inputenc}

\newtheorem{assumption}{Assumption}

\newtheorem{definition}{Definition}
\newtheorem{theorem}{Theorem}

\newtheorem{lemma}{Lemma}
\newtheorem{remark}{Remark}

\usepackage{newfloat}
\usepackage{listings}
\DeclareCaptionStyle{ruled}{labelfont=normalfont,labelsep=colon,strut=off} 
\floatstyle{ruled}
\newfloat{listing}{tb}{lst}{}
\floatname{listing}{Listing}
\newcommand{\norm}[1]{\left \lVert #1 \right\rVert }

\newcounter{relctr} 
\everydisplay\expandafter{\the\everydisplay\setcounter{relctr}{0}} 

\AtBeginDocument{}

\title{Regret Analysis of Policy Gradient Algorithm for Infinite Horizon Average Reward Markov Decision Processes }
\author{Qinbo Bai\equalcontrib, Washim Uddin Mondal\equalcontrib, Vaneet Aggarwal}
\affiliations{Purdue University, USA\\
\{bai113, wmondal, vaneet\}@purdue.edu}

\begin{document}

\maketitle

\begin{abstract}
    In this paper, we consider an infinite horizon average reward Markov Decision Process (MDP). Distinguishing itself from existing works within this context, our approach harnesses the power of the general policy gradient-based algorithm, liberating it from the constraints of assuming a linear MDP structure. We propose a policy gradient-based algorithm and show its global convergence property. We then prove that the proposed algorithm has $\tilde{\mathcal{O}}({T}^{3/4})$ regret. Remarkably, this paper marks a pioneering effort by presenting the first exploration into regret-bound computation for the general parameterized policy gradient algorithm in the context of average reward scenarios.
\end{abstract}


\section{Introduction}

Reinforcement Learning (RL) describes a class of problems where a learner repeatedly interacts with an unknown environment with the intention of maximizing the cumulative sum of rewards. This model has found its application in a wide array of areas, ranging from networking to transportation to epidemic control \citep{geng2020multi,al2019deeppool,ling2023cooperating}. RL problems are typically analysed via three distinct setups $-$ episodic, infinite horizon discounted reward, and infinite horizon average reward. Among these, the infinite horizon average reward setup holds particular significance in real-world applications (including those mentioned above) due to its alignment with many practical scenarios and its ability to capture essential long-term behaviors. However, scalable algorithms in this setup have not been widely studied. This paper provides an algorithm in the infinite horizon average reward setup with general parametrized policies which yields sub-linear regret guarantees. We would like to mention that this result is the first of its kind in the average reward setting. 

There are two major approaches to solving an RL problem. The first one, known as the model-based approach, involves constructing an estimate of the transition probabilities of the underlying Markov Decision Process (MDP). This estimate is subsequently leveraged to derive policies \citep{auer2008near,agrawal2017optimistic, ouyang2017learning, fruit2018efficient}. It is worth noting that model-based techniques encounter a significant challenge – these algorithms demand a substantial memory to house the model parameters. Consequently, their practical application is hindered when dealing with large state spaces. An alternative strategy is referred to as model-free algorithms. These methods either directly estimate the policy function or maintain an estimate of the $Q$ function, which are subsequently employed for policy generation \citep{mnih2015human, schulman2015trust, mnih2016asynchronous}. The advantage of these algorithms lies in their adaptability to handle large state spaces.

In the average reward MDP, which is the setting considered in our paper, one of the key performance indicators of an algorithm is the expected regret. It has been theoretically demonstrated in \citep{auer2008near} that the expected regret of any algorithm for a broad class of MDPs is lower bounded by $\Omega(\sqrt{T})$ where $T$ denotes the length of the time horizon. Many model-based algorithms, such as, \citep{auer2008near, agrawal2017optimistic} achieve this bound. Unfortunately, the above algorithms are designed to be applicable solely in the tabular setup. Recently, \citep{wei2021learning} proposed a model-based algorithm for the linear MDP setup that is shown to achieve the optimal regret bound. On the other hand, \citep{wei2020model} proposed a model-free $ Q$-estimation-based algorithm that achieves the optimal regret in the tabular setup. 
 
One way to extend algorithms beyond the tabular setting is via policy parameterization. Here, the policies are indexed by parameters (via, for example, neural networks), and the learning process is manifested by updating these parameters using some update rule (such as gradient descent). Such algorithms are referred to as policy gradient (PG) algorithms. Interestingly, the analysis of PG algorithms is typically restricted within the discounted reward setup. For example, \citep{agarwal2021theory} characterized the sample complexity of PG and Natural PG (NPG) with softmax and tabular parameterization. Sample complexity results for general parameterization are given by \citep{liu2020improved, NEURIPS2020_5f7695de}. However, the sub-linear regret analysis of a PG-based algorithm with general parameterization in the average reward setup, to the best of our knowledge, has not been studied in the literature. This paper aims to bridge this gap by addressing this crucial problem.

\subsection{Challenges and Contribution}

We propose a PG-based algorithm with general parameterization in the average reward setup and establish a sublinear regret of the proposed algorithm. In particular, within the class of ergodic MDPs, we first show that our PG-based algorithm achieves an average optimality error of  $\tilde{\mathcal{O}}(T^{-\frac{1}{4}})$. Utilizing this convergence result, we establish that the algorithm achieves a regret bound of $\Tilde{O}(T^{3/4})$. 

Despite the availability of sample complexity analysis of PG algorithms in the discounted reward setup, obtaining a sublinear regret bound for their average reward counterpart is quite difficult. This is because in the average reward case, the value function estimators, which are crucial to estimate the gradient, can become unbounded, unlike their discounted reward counterparts. Indeed, the sample complexity results in the discounted MDPs are often associated with a $\frac{1}{1-\gamma}$ factor (where $\gamma$ denotes the discount factor which is $1$ for the average reward case), indicating that a naive adaptation of these estimators will not work for the average reward setup. Also, discounted setups typically assume access to a simulator to generate unbiased value estimates. On the contrary, our paper deals with a single sample trajectory and does not assume the availability of a simulator. To obtain a good estimator of the gradient, we design an epoch-based algorithm where the length of each epoch is $H$. The algorithm estimates the value functions within a given epoch by sampling rewards of sub-trajectories of length $N$ that are at least $N$ distance apart. The separation between these sub-trajectories ensures that their reward samples are sufficiently independent. The key challenge of this paper is to bound a second-order term which is related to the variance of the estimated gradient and the true gradient. We show that by judiciously controlling the growth rate of $H$ and $N$ with $T$, it is possible to obtain a gradient estimator that has an asymptotically decreasing variance. 


\section{Related Works}

As discussed in the introduction, the reinforcement learning problem has been widely studied recently for infinite horizon discounted reward cases or the episodic setting. For example, \citep{NEURIPS2018_d3b1fb02} proposed the model-free UCB-Q learning and showed a $\mathcal{O}(\sqrt{T})$ regret in the episodic setting. In the discounted reward setting, \citep{NEURIPS2020_5f7695de} achieved  $\mathcal{O}(\epsilon^{-2})$ sample complexity for the softmax parametrization using the Natural Policy Gradient algorithm whereas \citep{mondal2023improved,fatkhullin2023stochastic} exhibited the same complexity for the general parameterization. However, the regret analysis or the global convergence of the average reward infinite horizon case is much less investigated. 

\begin{table*}[t]
    \centering
    \resizebox{0.8\textwidth}{!}{
        \begin{tabular}{|c|c|c|c|c|}
    	\hline
    	Algorithm & Regret & Ergodic & Model-free & Setting\\
    	\hline
            UCRL2 \citep{auer2008near} & $\Tilde{O}\bigg(DS\sqrt{AT}\bigg)$ & No & No & Tabular\\
    	\hline
            PSRL \citep{agrawal2017optimistic} & $\Tilde{O}\bigg(DS\sqrt{AT}\bigg)$ & No & No & Tabular\\
            \hline
            OPTIMISTIC Q-LEARNING \citep{wei2020model} & $\Tilde{O}\bigg(T^{2/3}\bigg)$ & No & Yes & Tabular\\
    	\hline
            MDP-OOMD \citep{wei2020model} & $\Tilde{O}\bigg(\sqrt{T}\bigg)$ & Yes & Yes & Tabular\\
    	\hline
            FOPO \citep{wei2021learning}\footnotemark[1] & $\Tilde{O}\bigg(\sqrt{T}\bigg)$ & No & No & Linear MDP\\
    	\hline
            OLSVI.FH \citep{wei2021learning} & $\Tilde{O}\bigg(T^{3/4}\bigg)$ & No & No & Linear MDP\\
    	\hline
            MDP-EXP2 \citep{wei2021learning} & $\Tilde{O}\bigg(\sqrt{T}\bigg)$ & Yes & No & Linear MDP\\
    	\hline
            This paper & $\Tilde{O}\bigg(T^{\frac{3}{4}}\bigg)$ & Yes & Yes & General parametrization\\
    	\hline\hline
            \textbf{Lower bound} \citep{auer2008near} &       $\Omega\bigg(\sqrt{DSAT}\bigg)$ & N/A & N/A & N/A\\
    	\hline
        \end{tabular}
    }
    \caption{ This table summarizes the different model-based and mode-free state-of-the-art algorithms available in the literature for average reward MDPs. We note that the proposed algorithm is the first paper to analyze the regret for average reward MDP with general parametrization.}
    \label{table2}
\end{table*}

For infinite horizon average reward MDPs, \citep{auer2008near} proposed a model-based Upper confidence Reinforcement learning (UCRL2) algorithm and established that it obeys a $\tilde{\mathcal{O}}(\sqrt{T})$ regret bound. \citep{agrawal2017optimistic} proposed posterior sampling-based approaches for average reward MDPs. \citep{wei2020model} proposed the optimistic-Q learning algorithm which connects the discounted reward and average reward setting together to show $\mathcal{O}(T^{3/4})$ regret in weakly communicating average reward case and another online mirror descent algorithm which achieves $\mathcal{O}(\sqrt{T})$ regret in the ergodic setting. For the linear MDP setting, \citep{wei2021learning} proposed three algorithms, including the MDP-EXP2 algorithm which achieves $O(\sqrt{T})$ regret under the ergodicity assumption. These works have been summarized in Table \ref{table2}. We note that the assumption of weakly communicating MDP is the minimum assumption needed to have sub-linear regret results. However, it is much more challenging to work with this assumption in the general parametrized setting because of the following reasons. Firstly, there is no guarantee that the state distribution will converge to the steady distribution exponentially fast which is the required property to show that the value functions are bounded by the mixing time. Secondly, it is unclear how to obtain an asymptotically unbiased estimate of the policy gradient. Thus, we assume ergodic MDP in this work following other works in the literature \citep{pesquerel2022imed, gong2020duality}. MDPs with constraints have also been recently studied for model-based \citep{agarwal2022regret,agarwal2022concave}, model-free tabular \citep{wei2022provably,chen2022learning}, and linear MDP setup \citep{ghosh2022achieving}. 

However, all of the above algorithms are designed for the tabular setting or the linear MDP assumption, and none of them uses a PG algorithm with the general parametrization setting. In this paper, we propose a PG algorithm for ergodic MDPs with general parametrization and analyze its regret. Our algorithm can, therefore, be applied beyond the tabular or linear MDP setting. 


\section{Formulation}

In this paper, we consider an infinite horizon reinforcement learning problem with an average reward criterion, which is modeled by the Markov Decision Process (MDP) written as a tuple $\mathcal{M}=(\mathcal{S},\mathcal{A},r, P,\rho)$ where $\mathcal{S}$ is the state space, $\mathcal{A}$ is the action space of size $A$, $r:\mathcal{S}\times\mathcal{A}\rightarrow [0,1]$ is the reward function,  $ P:\mathcal{S}\times\mathcal{A}\rightarrow \Delta^{|\mathcal{S}|}$ is the state transition function where $\Delta^{|\mathcal{S}|}$ denotes the probability simplex with dimension $|\mathcal{S}|$, and $\rho:\mathcal{S}\rightarrow[0,1]$ is the initial distribution of states. A policy $\pi:\mathcal{S}\rightarrow \Delta^{|\mathcal{A}|}$ decides the distribution of the action to be taken given the current state. For a given policy, $\pi$ we define the long-term average reward as follows.
\begin{equation}
    J_{\rho}^{\pi}\triangleq \lim\limits_{T\rightarrow \infty}\frac{1}{T}\mathbf{E}\bigg[\sum_{t=0}^{T-1}r(s_t,a_t)\bigg|s_0\sim \rho\bigg]
\end{equation}
where the expectation is taken over all state-action trajectories that are generated by following the action execution process, $a_t\sim\pi(\cdot|s_t)$ and the state transition rule, $s_{t+1}\sim P(\cdot|s_t, a_t)$, $\forall t\in\{0, 1, \cdots\}$. To simplify notations, we shall drop the dependence on $\rho$ whenever there is no confusion. We consider a parametrized class of policies, $\Pi$ whose each element is indexed by a $\mathrm{d}$-dimensional parameter, $\theta\in\Theta$ where $\Theta\subset\mathbb{R}^{\mathrm{d}}$. Our goal is to solve the following optimization problem.
\begin{equation}
    \max_{\theta\in\Theta} ~ J^{\pi_{\theta}} \triangleq J(\theta)
\end{equation}

A policy $\pi_{\theta}$ induces a transition function 
$P^{\pi_{\theta}}:\mathcal{S}\rightarrow \Delta^{|\mathcal{S}|}$ as  $P^{\pi_{\theta}}(s, s') = \sum_{a\in\mathcal{A}}P(s'|s,a)\pi_{\theta}(a|s)$, $\forall s, s'\in\mathcal{S}$. If $\mathcal{M}$ is such that for every policy $\pi$, the induced function, $P^{\pi}$
is irreducible, and aperiodic, then $\mathcal{M}$ is called ergodic.

\begin{assumption}
    \label{ass_1}
    The MDP $\mathcal{M}$ is ergodic.
\end{assumption}

\footnotetext[1]{FOPO is computationally inefficient.}

Ergodicity is commonly applied in the analysis of MDPs \citep{pesquerel2022imed, gong2020duality}. It is well known that if $\mathcal{M}$ is ergodic, then $\forall \theta\in\Theta$, there exists a unique stationary distribution, $d^{\pi_{\theta}}\in \Delta^{|\mathcal{S}|}$ defined as,
\begin{align}
    d^{\pi_{\theta}}(s) = \lim_{T\rightarrow \infty}\dfrac{1}{T}\left[\sum_{t=0}^{T-1} \mathrm{Pr}(s_t=s|s_0\sim \rho, \pi_{\theta})\right]
\end{align}

Note that under the assumption of ergodicity, $d^{\pi_{\theta}}$ is independent of the initial distribution, $\rho$, and satisfies $P^{\pi_{\theta}}d^{\pi_{\theta}}=d^{\pi_{\theta}}$. In this case, we can write the average reward as follows.
\begin{align}
    \label{eq_r_pi_theta}
    \begin{split}
        &J(\theta) = \mathbf{E}_{s\sim d^{\pi_{\theta}}, a\sim \pi_{\theta}(\cdot|s)}[r(s, a)] = (d^{\pi_{\theta}})^T r^{\pi_{\theta}} \\
        &\text{where}~r^{\pi_{\theta}}(s) \triangleq \sum_{a\in\mathcal{A}}r(s, a)\pi_{\theta}(a|s), ~\forall s\in \mathcal{S}
    \end{split}
\end{align}

Hence, the average reward $J(\theta)$ is also independent of the initial distribution, $\rho$. Furthermore, $\forall \theta\in\Theta$, there exist a function $Q^{\pi_{\theta}}: \mathcal{S}\times \mathcal{A}\rightarrow \mathbb{R}$ such that the following Bellman equation is satisfied $\forall (s, a)\in\mathcal{S}\times\mathcal{A}$.
\begin{equation}
    \label{eq_bellman}
    Q^{\pi_{\theta}}(s,a)=r(s,a)-J(\theta)+\mathbf{E}_{s'\sim P(\cdot|s, a)}\left[V^{\pi_{\theta}}(s')\right]
\end{equation}
where the state value function, $V^{\pi_{\theta}}:\mathcal{S}\rightarrow \mathbb{R}$ is defined as,
\begin{align}
    \label{eq_V_Q}
    V^{\pi_{\theta}}(s) = \sum_{a\in\mathcal{A}}\pi_{\theta} (a|s)Q^{\pi_{\theta}}(s, a), ~\forall s\in\mathcal{S}
\end{align}

Note that if $(\ref{eq_bellman})$ is satisfied by $Q^{\pi_{\theta}}$, then it is also satisfied by $Q^{\pi{\theta}}+c$ for any arbitrary constant, $c$. To define these functions uniquely, we assume that $\sum_{s\in\mathcal{S}}d^{\pi_{\theta}}(s)V^{\pi_{\theta}}(s)=0$. In this case, $V^{\pi_{\theta}}(s)$ can be written as follows $\forall s\in\mathcal{S}$.
\begin{align}
    \label{def_v_pi_theta_s}
    \begin{split}
        V^{\pi_{\theta}}(s) &= \sum_{t=0}^{\infty} \sum_{s'\in\mathcal{S}}\left[(P^{\pi_{\theta}})^t(s, s') - d^{\pi_{\theta}}(s')\right]r^{\pi_{\theta}}(s')\\
        &=\mathbf{E}_{\theta}\left[\sum_{t=0}^{\infty}r(s_t,a_t)-J(\theta)\bigg\vert s_0=s\right]
    \end{split}
\end{align}
where $\mathbf{E}_{\theta}[\cdot]$ denotes expectation over all trajectories induced by the policy $\pi_{\theta}$. Similarly, $\forall (s, a)\in\mathcal{S}\times \mathcal{A}$, $Q^{\pi_{\theta}}(s, a)$ can be uniquely written as,
\begin{equation}
    Q^{\pi_{\theta}}(s,a)=\mathbf{E}_{\theta}\left[\sum_{t=0}^{\infty}r(s_t,a_t)-J(\theta)\bigg\vert s_0=s,a_0=a\right]
\end{equation}

Additionally, we define the advantage function $A^{\pi_{\theta}}:\mathcal{S}\times \mathcal{A}\rightarrow \mathbb{R}$ such that $\forall (s, a)\in\mathcal{S}\times\mathcal{A}$,
\begin{align}
    A^{\pi_{\theta}}(s, a) \triangleq Q^{\pi_{\theta}}(s, a) - V^{\pi_{\theta}}(s)
\end{align}

Ergodicity also implies the existence of a finite mixing time. In particular, if $\mathcal{M}$ is ergodic, then the mixing time is defined as follows.
\begin{definition}
The mixing time of an MDP $\mathcal{M}$ with respect to a policy parameter $\theta$ is defined as,
\begin{align}
    t_{\mathrm{mix}}^{\theta}\triangleq \min\left\lbrace t\geq 1\big| \norm{(P^{\pi_{\theta}})^t(s, \cdot) - d^{\pi_\theta}}\leq \dfrac{1}{4}, \forall s\in\mathcal{S}\right\rbrace 
\end{align} 

We also define $t_{\mathrm{mix}}\triangleq \sup_{\theta\in\Theta} t^{\theta}_{\mathrm{mix}} $ as the the overall mixing time. In this paper,  $t_{\mathrm{mix}}$ is finite due to ergodicity.
\end{definition}

Mixing time is a measure of how fast the MDP reaches close to its stationary distribution if the same policy is kept on being executed repeatedly. We also define the hitting time as follows.
\begin{definition}
The hitting time of an MDP $\mathcal{M}$ with respect to a policy parameter, $\theta$ is defined as,
\begin{align}
    t_{\mathrm{hit}}^{\theta}\triangleq  \max_{s\in\mathcal{S}} \dfrac{1}{d^{\pi_{\theta}}(s)}
\end{align} 

We also define $t_{\mathrm{hit}}\triangleq \sup_{\theta\in\Theta} t^{\theta}_{\mathrm{hit}} $ as the the overall hitting time. In this paper, $t_{\mathrm{hit}}$ is finite due to ergodicity.
\end{definition}

Let, $J^* \triangleq \sup_{\boldsymbol{\theta}\in\Theta} J(\theta)$. For a given MDP $\mathcal{M}$ and a time horizon $T$, the regret of an algorithm $\mathbb{A}$ is defined as follows.
\begin{align}
    \mathrm{Reg}_T(\mathbb{A}, \mathcal{M}) \triangleq \sum_{t=0}^{T-1} \left(J^*-r(s_t, a_t)\right)
\end{align}
where the action, $a_t$, $t\in\{0, 1, \cdots \}$ is chosen by following the algorithm, $\mathbb{A}$ based on the trajectory up to time, $t$, and the state, $s_{t+1}$ is obtained by following the state transition function, $P$. Wherever there is no confusion, we shall simplify the notation of regret to $\mathrm{Reg}_{T}$. The goal of maximizing $J(\cdot)$ can be accomplished by designing an algorithm that minimizes the regret.


\section{Algorithm}

In this section, we discuss a policy-gradient-based algorithm in the average reward RL settings. For simplicity, we assume that the set of all policy parameters is $\Theta = \mathbb{R}^{\mathrm{d}}$. The standard policy gradient algorithm iterates the policy parameter $\theta$ as follows $\forall k\in\{1,2,\cdots\}$ starting with an initial guess $\theta_1$.
\begin{equation}
    \theta_{k+1}=\theta_k+\alpha\nabla_\theta J(\theta_k)
\end{equation}
where $\alpha$ is the parameter learning rate. The following result is well-known in the literature \citep{sutton1999policy}.

\begin{lemma}
    \label{lemma_grad_compute}
    The gradient of the long-term average reward can be expressed as follows.
    \begin{align}
        &\nabla_{\theta} J(\theta)=\mathbf{E}_{s\sim d^{\pi_{\theta}},a\sim\pi_{\theta}(\cdot|s)}\bigg[ A^{\pi_{\theta}}(s,a)\nabla_{\theta}\log\pi_{\theta}(a|s)\bigg]
    \end{align}
\end{lemma}
    	
Typically we have access neither to $P$, the state transition function to compute the required expectation nor to the functions $V^{\pi_{\theta}}$, $Q^{\pi_{\theta}}$. In the absence of this knowledge, computation of gradient, therefore, becomes a difficult job. In the subsequent discussion, we shall demonstrate how the gradient can be estimated using sampled trajectories. Our policy gradient-based algorithm is described in Algorithm \ref{alg:PG_MAG}.

The algorithm proceeds in multiple epochs with the length of each epoch being $H=16t_{\mathrm{hit}}t_{\mathrm{mix}}\sqrt{T}(\log T)^2$. Observe that the algorithm is assumed to be aware of $T$. This assumption can be easily relaxed invoking the well-known doubling trick \citep{lattimore2020bandit}. 
We also assume that the values of $t_{\mathrm{mix}}$, and $t_{\mathrm{hit}}$ are known to the algorithm. Similar presumptions have been used in the previous literature \citep{wei2020model}. In the $k$th epoch, the algorithm generates a trajectory of length $H$, denoted as $\mathcal{T}_k=\{(s_{t}, a_{t})\}_{t=(k-1)H}^{kH-1}$, by following the policy $\pi_{\theta_k}$.
We utilise the policy parameter $\theta_k$ and the trajectory $\mathcal{T}_k$ in Algorithm \ref{alg:estQ} to compute the estimates $\hat{V}^{\pi_{\theta_k}}(s)$, and $\hat{Q}^{\pi_{\theta_k}}(s, a)$ for a given state-action pair $(s, a)$. The algorithm searches the trajectory $\mathcal{T}_k$ to locate disjoint sub-trajectories of length $N=4t_{\mathrm{mix}}(\log T)$ that start with the given state $s$ and are at least $N$ distance apart. Let $i$ be the number of such sub-trajectories and the sum of rewards in the $j$th such sub-trajectory be $y_j$. Then $\hat{V}^{\pi_{\theta_k}}(s)$ is computed as,
\begin{align}
    \label{eq_V_estimate}
    \hat{V}^{\pi_{\theta_k}}(s)=\dfrac{1}{i}\sum_{j=1}^i y_j
\end{align}

The sub-trajectories are kept at least $N$ distance apart to ensure that the samples $\{y_j\}_{j=1}^i$ are fairly independent. The estimate $\hat{Q}^{\pi_{\theta}}(s, a)$, on the other hand, is given as, 
\begin{align}
    \label{eq_reward_tracking}
    \begin{split}
        \hat{Q}^{\pi_{\theta}}(s, a) = \dfrac{1}{\pi_{\theta_k}(a|s)}\left[\dfrac{1}{i}\sum_{j=1}^i y_j\mathrm{1}(a_{\tau_j}=a)\right]
    \end{split}
\end{align}
where $\tau_j$ is the starting time of the $j$th chosen sub-trajectory. Finally, the advantage value is estimated as,
\begin{align}
    \hat{A}^{\pi_{\theta_k}}(s, a) = \hat{Q}^{\pi_{\theta_k}}(s, a) - \hat{V}^{\pi_{\theta_k}}(s)
\end{align}

This allows us to compute an estimate of the policy gradient as follows.
\begin{align}
    \label{eq_grad_estimate}
    \omega_k\triangleq\hat{\nabla}_{\theta} J(\theta_k) = \dfrac{1}{H}\sum_{t=t_k}^{t_{k+1}-1}\hat{A}^{\pi_{\theta}}(s_{t}, a_{t})\nabla_{\theta}\log \pi_{\theta_k}(a_{t}|s_{t})
\end{align}
where $t_k=(k-1)H$ is the starting time of the $k$th epoch. The policy parameters are updated via \eqref{udpates_algorotihm}. In the following lemma, we show that $\hat{A}^{\pi_{\theta_k}}(s, a)$ is a good estimator of $A^{\pi_{\theta_k}}(s, a)$.

\begin{lemma}
    \label{lemma_good_estimator}
    The following inequalities hold $\forall k$, $\forall (s, a)$ and sufficiently large $T$.
    \begin{align}
        \label{eq_good_estimator}
        \begin{split}
            \mathbf{E}&\bigg[\bigg(\hat{A}^{\pi_{\theta_k}}(s, a)-A^{\pi_{\theta_k}}(s, a)\bigg)^2\bigg]\\
            &\leq \mathcal{O}\left(\dfrac{t_{\mathrm{hit}}N^3\log T}{H\pi_{\theta_k}(a|s)}\right) =\mathcal{O}\left(\dfrac{t_{\mathrm{mix}}^2(\log T)^2}{\sqrt{T}\pi_{\theta_k}(a|s)}\right)
        \end{split}
    \end{align}
\end{lemma}

Lemma \ref{lemma_good_estimator} establishes that the $L_2$ error of our proposed estimator can be bounded above as $\Tilde{\mathcal{O}}(1/\sqrt{T})$. As we shall see later, this result can be used to bound the estimation error of the gradient. It is worthwhile to point out that $\hat{V}^{\pi_{\theta_k}}(s)$ and $\hat{Q}^{\pi_{\theta_k}}(s, a)$ defined in $(\ref{eq_V_estimate})$, $(\ref{eq_reward_tracking})$ respectively, may not themselves be good estimators of their target quantities although their difference is one. We would also like to mention that our Algorithm \ref{alg:estQ} is inspired by Algorithm 2 of \citep{wei2020model}. The main difference is that we take the episode length to be $H=\tilde{\mathcal{O}}(\sqrt{T})$ while in \citep{wei2020model}, it was chosen to be $\tilde{\mathcal{O}}(1)$. This extra $\sqrt{T}$ factor makes the estimation error a decreasing function of $T$. 

\begin{algorithm}[t]
    \caption{Parameterized Policy Gradient}
    \label{alg:PG_MAG}
    \begin{algorithmic}[1]
        \STATE \textbf{Input:} Initial parameter $\theta_1$, learning rate $\alpha$,  initial state $s_0 \sim \rho(\cdot)$, episode length $H$ \vspace{0.1cm}
        \STATE $K=T/H$
        
	\FOR{$k\in\{1, \cdots, K\}$}
            \STATE $\mathcal{T}_k\gets \phi$
            
            \FOR{$t\in\{(k-1)H, \cdots, kH-1\}$}
                \STATE Execute $a_t\sim \pi_{\theta_k}(\cdot|s_t)$, receive reward $r(s_t,a_t) $ and observe $s_{t+1}$
                \STATE $\mathcal{T}_k\gets \mathcal{T}_k\cup \{(s_t, a_t)\}$
            \ENDFOR	
            
            \FOR{$t\in\{(k-1)H, \cdots, kH-1\}$}
                \STATE Using Algorithm \ref{alg:estQ}, and $\mathcal{T}_k$, compute $\hat{A}^{\pi_{\theta_k}}(s_t, a_t)$
            \ENDFOR
            \vspace{0.1cm}
      
            \STATE Using \eqref{eq_grad_estimate}, compute  $\omega_k$ 
		\STATE Update parameters as
		\begin{equation}
                \label{udpates_algorotihm}
		    \theta_{k+1}=\theta_k+\alpha\omega_k
		\end{equation}
        \ENDFOR
    \end{algorithmic}
\end{algorithm}

\begin{algorithm}[t]
    \caption{Advantage Estimation}
    \label{alg:estQ}
    \begin{algorithmic}[1]
        \STATE \textbf{Input:} Trajectory $(s_{t_1}, a_{t_1},\ldots, s_{t_2}, a_{t_2})$, state $s$, action $a$, and policy parameter $\theta$
        \STATE \textbf{Initialize:} $i \leftarrow 0$, $\tau\leftarrow t_1$
	\STATE \textbf{Define:} $N=4t_{\mathrm{mix}}\log_2T$.
	\vspace{0.1cm}
	\WHILE{$\tau\leq t_2-N$}
		\IF{$s_{\tau}=s$}
			\STATE $i\leftarrow i+1$.
			\STATE $\tau_i\gets \tau$
			\STATE $y_i=\sum_{t=\tau}^{\tau+N-1}r(s_t, a_t)$.
			\STATE $\tau\leftarrow\tau+2N$.	
		\ELSE
                \STATE {$\tau\leftarrow\tau+1$.}
            \ENDIF
	\ENDWHILE
        \vspace{0.1cm}
        \IF{$i>0$}
            \STATE $\hat{V}(s)=\dfrac{1}{i}\sum_{j=1}^i y_j$,
            \STATE $\hat{Q}(s, a) = \dfrac{1}{\pi_{\theta}(a|s)}\left[\dfrac{1}{i}\sum_{j=1}^i y_j\mathrm{1}(a_{\tau_j}=a)\right]$
        \ELSE
            \STATE $\hat{V}(s)=0$, $\hat{Q}(s, a) = 0$
        \ENDIF
	\STATE \textbf{return}  $\hat{Q}(s, a)-\hat{V}(s)$ 
    \end{algorithmic}
\end{algorithm}


\section{Global Convergence Analysis}

In this section, we show that our proposed Algorithm \ref{alg:PG_MAG} converges globally. This essentially means that the parameters $\{\theta_k\}_{k=1}^{\infty}$ are such that the sequence $\{J(\theta_k)\}_{k=1}^{\infty}$, in certain sense, approaches the optimal average reward, $J^*$. Such convergence will be later useful in bounding the regret of our algorithm. Before delving into the analysis, we would like to first point out a few assumptions that are needed to establish the results.
\begin{assumption}
    \label{ass_score}
    The log-likelihood function is $G$-Lipschitz and $B$-smooth. Formally, $\forall \theta, \theta_1,\theta_2 \in\Theta,\forall (s,a)\in\mathcal{S}\times\mathcal{A}$
    \begin{equation}
	\begin{aligned}
            &\Vert\nabla_{\theta}\log\pi_\theta(a\vert s)\Vert\leq G\quad\forall \theta\in\Theta,\forall (s,a)\in\mathcal{S}\times\mathcal{A}\\
            &\Vert \nabla_{\theta}\log\pi_{\theta_1}(a\vert s)-\nabla_\theta\log\pi_{\theta_2}(a\vert s)\Vert\leq B\Vert \theta_1-\theta_2\Vert\quad
	\end{aligned}
    \end{equation}
\end{assumption}

\begin{remark}
    The Lipschitz and smoothness properties for the log-likelihood are quite common in the field of policy gradient algorithm \citep{Alekh2020, Mengdi2021, liu2020improved}. Such properties can also be verified for simple parameterization such as Gaussian policy.   
\end{remark}

One can immediately see that by combining Assumption \ref{ass_score} with Lemma \ref{lemma_good_estimator} and using the definition of the gradient estimator as given in $(\ref{eq_grad_estimate})$, we arrive at the following important result.

\begin{lemma}
    \label{lemma_grad_est_bias}
    The following relation holds $\forall k$.
    \begin{align}
        \mathbf{E}\left[\norm{\omega_k-\nabla_{\theta}J(\theta_k)}^2\right]\leq \mathcal{O}\left(\dfrac{AG^2t_{\mathrm{mix}}^2(\log T)^2}{\sqrt{T}}\right)
    \end{align}
\end{lemma}

Lemma \ref{lemma_grad_est_bias} claims that the error in estimating the gradient can be bounded above as $\tilde{\mathcal{O}}(1/\sqrt{T})$. This result will be used in proving the global convergence of our algorithm.

\begin{assumption}
    \label{ass_transfer_error}
    Define the transferred function approximation error
    \begin{align}
        \label{eq:transfer_error}
        \begin{split}
            L_{d_\rho^{\pi^*},\pi^*}(\omega^*_\theta,\theta &) =\mathbf{E}_{s\sim d_\rho^{\pi^*}}\mathbf{E}_{a\sim\pi^*(\cdot\vert s)}\bigg[\\
            &\bigg(\nabla_\theta\log\pi_{\theta}(a\vert s)\cdot\omega^*_{\theta}-A^{\pi_\theta}(s,a)\bigg)^2\bigg]
	\end{split}
    \end{align}
    where $\pi^*$ is the optimal policy and $\omega^*_{\theta}$ is given as
    \begin{align}
        \label{eq:NPG_direction}
	\begin{split}
            \omega^*_{\theta}=\arg&\min_{\omega\in\mathbb{R}^{\mathrm{d}}}~\mathbf{E}_{s\sim d_\rho^{\pi_{\theta}}}\mathbf{E}_{a\sim\pi_{\theta}(\cdot\vert s)}\bigg[\\
            &\bigg(\nabla_\theta\log\pi_{\theta}(a\vert s)\cdot\omega-A^{\pi_{\theta}}(s,a)\bigg)^2\bigg]
	\end{split}
    \end{align}
    We assume that the error satisfies $L_{d_{\rho}^{\pi^*},\pi^*}(\omega^*_{\theta},\theta)\leq \epsilon_{\mathrm{bias}}$ for any $\theta\in\Theta$ where $\epsilon_{\mathrm{bias}}$ is a positive constant.
\end{assumption}
 
\begin{remark}
    The transferred function approximation error, defined by  \eqref{eq:transfer_error} and \eqref{eq:NPG_direction}, quantifies the expressivity of the policy class in consideration. It has been shown that the softmax parameterization \citep{agarwal2021theory} or linear MDP structure \citep{Chi2019} admits $\epsilon_{\mathrm{bias}}=0$. When parameterized by the restricted policy class that does not contain all the policies, $\epsilon_{\mathrm{bias}}$ turns out to be strictly positive. However, for a rich neural network parameterization, the $\epsilon_{bias}$ is small \citep{Lingxiao2019}. A similar assumption has been adopted in \citep{liu2020improved} and \citep{agarwal2021theory}. 
\end{remark}

\begin{remark}
    It is to be mentioned that $\omega^*_\theta$ defined in \eqref{eq:NPG_direction} can be alternatively written as,
    \begin{align*}
        \omega^*_{\theta} = F(\theta)^{\dagger} \mathbf{E}_{s\sim d^{\pi_{\theta}}}\mathbf{E}_{a\sim\pi_{\theta}(\cdot\vert s)}\left[\nabla_{\theta}\log\pi_{\theta}(a|s)A^{\pi_{\theta}}(s, a)\right]
    \end{align*}
    where $\dagger$ symbolizes the Moore-Penrose pseudoinverse operation and $F(\theta)$ is the Fisher information matrix as defined below.
    \begin{align}
        \begin{split}
            F(\theta) = \mathbf{E}_{s\sim d^{\pi_{\theta}}}&\mathbf{E}_{a\sim\pi_{\theta}(\cdot\vert s)}\left[\right.\\
            &\left.\nabla_{\theta}\log\pi_{\theta}(a|s)(\nabla_{\theta}\log\pi_{\theta}(a|s))^T\right]
        \end{split}
    \end{align}
\end{remark}

\begin{assumption}
    \label{ass_4}
    There exists a constant $\mu_F>0$ such that $F(\theta)-\mu_F I_{\mathrm{d}}$ is positive semidefinite where $I_{\mathrm{d}}$ denotes an identity matrix.
\end{assumption}

Assumption \ref{ass_4} is also commonly used in the policy gradient analysis \citep{liu2020improved}. This is satisfied by the Gaussian policy with a linearly parameterized mean.

In the discounted reward setup, one key result is the performance difference lemma. In the averaged reward setting, this is derived as stated below.
\begin{lemma}
    \label{lem_performance_diff}
    The difference in the performance for  any policies $\pi_\theta$ and $\pi_{\theta'}$is bounded as follows
    \begin{equation}
        J(\theta)-J(\theta')= \mathbf{E}_{s\sim d^{\pi_\theta}}\mathbf{E}_{a\sim\pi_\theta(\cdot\vert s)}\big[A^{\pi_{\theta'}}(s,a)\big]
    \end{equation}
\end{lemma}

Using Lemma \ref{lem_performance_diff}, we present a general framework for convergence analysis of the policy gradient algorithm in the averaged reward case as dictated below. This is inspired by the convergence analysis of \citep{liu2020improved} for the discounted reward MDPs. 

\begin{lemma}
    \label{lem_framework} 
    Suppose a general gradient ascent algorithm updates the policy parameter in the following way.
    \begin{equation}
	\theta_{k+1}=\theta_k+\alpha\omega_k
    \end{equation}
    When Assumptions \ref{ass_score}, \ref{ass_transfer_error}, and \ref{ass_4} hold, we have the following inequality for any $K$.
    \begin{equation}
        \label{eq:general_bound}
	\begin{split}
            &J^{*}-\frac{1}{K}\sum_{k=1}^{K}J(\theta_k)\leq \sqrt{\epsilon_{\mathrm{bias}}}+\frac{G}{K}\sum_{k}^{K}\Vert(\omega_k-\omega^*_k)\Vert\\
            &+\frac{B\alpha}{2K}\sum_{k=1}^{K}\Vert\omega_k\Vert^2+\frac{1}{\alpha K}\mathbf{E}_{s\sim d^{\pi^*}}[KL(\pi^*(\cdot\vert s)\Vert\pi_{\theta_1}(\cdot\vert s))]		
        \end{split}
    \end{equation}
    where $\omega^*_k:=\omega^*_{\theta_k}$ and $\omega^*_{\theta_k}$ is defined in \eqref{eq:NPG_direction}, $J^*=J(\theta^*)$, and $\pi^*=\pi_{\theta^*}$ where $\theta^*$ is the optimal parameter.
\end{lemma}

Lemma \ref{lem_framework} bounds the optimality error of any gradient ascent algorithm as a function of intermediate gradient norms. Note the presence of $\epsilon_{\mathrm{bias}}$ in the upper bound. Clearly, for a severely restricted policy class where $\epsilon_{\mathrm{bias}}$ is significant, the optimality bound becomes poor. Consider the expectation of the second term in \eqref{eq:general_bound}. Note that,
\begin{equation}
    \label{eq_second_term_bound}
    \begin{split}
        &\bigg(\frac{1}{K}\sum_{k=1}^{K}\mathbf{E}\Vert\omega_k-\omega^*_k\Vert\bigg)^2\leq \frac{1}{K}\sum_{k=1}^{K}\mathbf{E}\bigg[\Vert\omega_k-\omega^*_k\Vert^2\bigg]\\
        &=\frac{1}{K}\sum_{k=1}^{K}\mathbf{E}\bigg[\Vert\omega_k-F(\theta_k)^\dagger\nabla_\theta J(\theta_k)\Vert^2\bigg]\\
        &\leq \frac{2}{K}\sum_{k=1}^{K}\mathbf{E}\bigg[\Vert\omega_k-\nabla_{\theta}J(\theta_k)\Vert^2\bigg]\\
        &\hspace{1cm}+\frac{2}{K}\sum_{k=1}^{K}\mathbf{E}\bigg[\Vert \nabla_{\theta} J(\theta_k)- F(\theta_k)^\dagger\nabla_\theta J(\theta_k)\Vert^2\bigg]\\
        &\overset{(a)}{\leq} \frac{2}{K}\sum_{k=1}^{K}\mathbf{E}\bigg[\Vert \omega_k-\nabla_{\theta}J(\theta_k)\Vert^2\bigg]\\
        &\hspace{1cm}+\frac{2}{K}\sum_{k=1}^{K}\left(1+\dfrac{1}{\mu_F^2}\right)\mathbf{E}\bigg[\Vert \nabla_\theta J(\theta_k)\Vert^2\bigg]
    \end{split}
\end{equation}
where $(a)$ uses Assumption \ref{ass_4}. The expectation of the third term in \eqref{eq:general_bound} can be bounded as follows.
 
\begin{equation}
    \label{eq_third_term_bound}
    \begin{split}
        \dfrac{1}{K}\sum_{k=1}^{K}\mathbf{E}\bigg[\Vert\omega_k\Vert^2\bigg]&\leq \dfrac{1}{K}\sum_{k=1}^{K}\mathbf{E}\bigg[\Vert \omega_k-\nabla_{\theta}J(\theta_k)\Vert^2\bigg] \\
        &+ \dfrac{1}{K}\sum_{k=1}^{K}\mathbf{E}\left[\norm{\nabla_{\theta } J(\theta_k)}^2\right] 
    \end{split}
\end{equation}

In both \eqref{eq_second_term_bound}, \eqref{eq_third_term_bound}, the terms related to $\norm{\omega_k-\nabla_{\theta}J(\theta_k)}^2$ are bounded by Lemma \ref{lemma_grad_est_bias}. To bound the term $\norm{\nabla_{\theta}J(\theta_k)}^2$, we use the following lemma.
 
\begin{lemma} 
    \label{lemma:41ss}
    Let $J(\cdot)$ be $L$-smooth and $\alpha = \frac{1}{4L}$. Then the following inequality holds.
    \begin{align*}
        \begin{split}
           \dfrac{1}{K}\sum_{k=1}^{K} \norm{ \nabla J(\theta_k) }^2 \leq &\dfrac{16L}{K}+ 
           \dfrac{16}{3K}\sum_{k=1}^{K}\Vert\nabla J(\theta_k) - \omega_k\Vert^2 
        \end{split}
    \end{align*}
\end{lemma}

Using Lemma \ref{lemma_grad_est_bias}, we obtain the following inequality.
\begin{align}
    \label{eq_33}
    \dfrac{1}{K}\sum_{k=1}^{K} \norm{ \nabla_{\theta} J(\theta_k) }^2 \leq \mathcal{\tilde{O}}\left(\dfrac{AG^2t_{\mathrm{mix}}^2+Lt_{\mathrm{mix}}t_{\mathrm{hit}}}{\sqrt{T}}\right)
\end{align}

Applying Lemma \ref{lemma_grad_est_bias} and \eqref{eq_33} in \eqref{eq_third_term_bound}, we finally get,
\begin{align}
    \label{eq_34}
    \dfrac{1}{K}\sum_{k=1}^{K}\mathbf{E}\bigg[\Vert\omega_k\Vert^2\bigg]&\leq \mathcal{\tilde{O}}\left(\dfrac{AG^2t_{\mathrm{mix}}^2+Lt_{\mathrm{mix}}t_{\mathrm{hit}}}{\sqrt{T}}\right)
\end{align}

Similarly, using $(\ref{eq_second_term_bound})$, we deduce the following.
\begin{align}
    \label{eq_35}
    \frac{1}{K}\sum_{k=1}^{K}\mathbf{E}\Vert\omega_k-\omega^*_k\Vert  \leq \mathcal{\tilde{O}}\left(\dfrac{\sqrt{A}Gt_{\mathrm{mix}}+\sqrt{Lt_{\mathrm{mix}}t_{\mathrm{hit}}}}{T^{\frac{1}{4}}(1+1/\mu_F)^{-1}}\right)
\end{align}

Inequalities \eqref{eq_34} and \eqref{eq_35} lead to the following result.

\begin{theorem}
    \label{theorem_1_statement}
    Let $\{\theta_k\}_{k=1}^{K}$ be defined as in Lemma \ref{lem_framework}. If assumptions \ref{ass_1}, \ref{ass_score}, \ref{ass_transfer_error},  \ref{ass_4} hold,  $J(\cdot)$ is $L$-smooth and $\alpha=\frac{1}{4L}$, then the following inequality holds for $K=T/H$ where $T$ is sufficiently large and $H=16t_{\mathrm{mix}}t_{\mathrm{hit}}\sqrt{T}(\log_2 T)^2$.
    \begin{align}
        \label{thm_final_convergence}
        \begin{split}
            &J^{*}-\frac{1}{K}\sum_{k=1}^{K}\mathbf{E}\left[J(\theta_k)\right]\leq \mathcal{\tilde{O}}\left(\dfrac{ABG^2t_{\mathrm{mix}}^2+BLt_{\mathrm{mix}}t_{\mathrm{hit}}}{L\sqrt{T}}\right) \\
            &+ \mathcal{\tilde{O}}\left(\dfrac{\sqrt{A}G^2t_{\mathrm{mix}}+G\sqrt{Lt_{\mathrm{mix}}t_{\mathrm{hit}}}}{T^{\frac{1}{4}}(1+1/\mu_F)^{-1}}\right)+  \sqrt{\epsilon_{\mathrm{bias}}}\\
            &+\mathcal{\tilde{O}}\left(\dfrac{Lt_{\mathrm{mix}}t_{\mathrm{hit}}}{\sqrt{T}}\mathbf{E}_{s\sim d^{\pi^*}}[KL(\pi^*(\cdot\vert s)\Vert\pi_{\theta_1}(\cdot\vert s))]\right)
        \end{split}
    \end{align}
\end{theorem}

Theorem \ref{theorem_1_statement} dictates that the sequence $\{J(\theta_k)\}_{k=1}^{K}$ generated by Algorithm \ref{alg:PG_MAG} converges to $J^*$ with a convergence rate of $\mathcal{O}(T^{-\frac{1}{4}}+\sqrt{\epsilon_{\mathrm{bias}}})$. Alternatively, one can say that in order to achieve an optimality error of $\epsilon+\sqrt{\epsilon_{\mathrm{bias}}}$, it is sufficient to choose $T=\mathcal{O}\left(\epsilon^{-4}\right)$. It matches the state-of-the-art sample complexity bound of the policy gradient algorithm with general parameterization in the discounted reward setup \citep{liu2020improved}.


\section{Regret Analysis}

In this section, we demonstrate how the convergence analysis in the previous section can be used to bind the expected regret of our proposed algorithm. Note that the regret can be decomposed as follows.
\begin{align}
    \label{reg_decompose}
    \begin{split}
        &\mathrm{Reg}_T = \sum_{t=0}^{T-1} \left(J^* - r(s_t, a_t)\right)\\
        &=H\sum_{k=1}^{K}\left(J^*-J({\theta_k})\right)+\sum_{k=1}^{K}\sum_{t\in\mathcal{I}_k} \left(J(\theta_k)-r(s_t, a_t)\right)
    \end{split}
\end{align}
where $\mathcal{I}_k\triangleq \{(k-1)H, \cdots, kH-1\}$. Note that the expectation of the first term in $(\ref{reg_decompose})$ can be bounded using Theorem \ref{thm_final_convergence}. The expectation of the second term can be expressed as follows,
\begin{align}
    \label{eq_38}
    \begin{split}
        &\mathbf{E}\left[\sum_{k=1}^{K}\sum_{t\in\mathcal{I}_k} \left(J(\theta_k)-r(s_t, a_t)\right)\right]\\
        &\overset{(a)}{=}\mathbf{E}\left[\sum_{k=1}^{K}\sum_{t\in\mathcal{I}_k} \mathbf{E}_{s'\sim P(\cdot|s_t, a_t)}[V^{\pi_{\theta_k}}(s')]-Q^{\pi_{\theta_k}}(s_t, a_t)\right]\\
        &\overset{(b)}{=}\mathbf{E}\left[\sum_{k=}^{K}\sum_{t\in\mathcal{I}_k} V^{\pi_{\theta_k}}(s_{t+1})-V^{\pi_{\theta_k}}(s_t)\right]\\
        &=\mathbf{E}\left[\sum_{k=1}^{K} V^{\pi_{\theta_k}}(s_{kH})-V^{\pi_{\theta_k}}(s_{(k-1)H})\right]\\
        &=\underbrace{\mathbf{E}\left[\sum_{k=1}^{K-1} V^{\pi_{\theta_{k+1}}}(s_{kH})-V^{\pi_{\theta_k}}(s_{kH})\right]}_{\triangleq P}\\
        &+\underbrace{\mathbf{E}\left[ V^{\pi_{\theta_K}}(s_{T})-V^{\pi_{\theta_0}}(s_{0})\right]}_{\triangleq Q}
    \end{split}
\end{align}
where $(a)$ follows from Bellman equation and $(b)$ utilises the following facts: $\mathbf{E}[V^{\pi_{\theta_k}}(s_{t+1})] = \mathbf{E}_{s'\sim P(\cdot|s_t, a_t)}[V^{\pi_{\theta_k}}(s')]$ and $\mathbf{E}[V^{\pi_{\theta_k}}(s_{t})]=\mathbf{E}[Q^{\pi_{\theta_k}}(s_t, a_t)]$. The term, $P$ in \eqref{eq_38} can be bounded using Lemma \ref{lemma_last} (stated below). Moreover, the term $Q$ can be upper bounded as $\mathcal{O}(t_{\mathrm{mix}})$ as clarified in the appendix.

\begin{lemma}
    \label{lemma_last}
    If assumptions \ref{ass_1} and \ref{ass_score} hold, then for $K=T/H$ where $H=16t_{\mathrm{mix}}t_{\mathrm{hit}}\sqrt{T}(\log_2 T)^2$, the following inequalities are true $\forall k$, $\forall (s, a)\in\mathcal{S}\times \mathcal{A}$ and sufficiently large $T$.
    \begin{align*}
        &(a) ~|\pi_{\theta_{k+1}}(a|s)-\pi_{\theta_{k}}(a|s)|\leq G\pi_{\bar{\theta}_k}(a|s)\norm{\theta_{k+1}-\theta_k}\\
        &(b) \sum_{k=1}^{K}\mathbf{E}|J(\theta_{k+1})-J(\theta_k)|\leq  \mathcal{\tilde{O}}\left(\dfrac{\alpha G C}{t_{\mathrm{hit}}}T^{\frac{1}{4}}\right)\\
        &(c) \sum_{k=1}^K\mathbf{E}|V^{\pi_{\theta_{k+1}}}(s_k) - V^{\pi_{\theta_{k}}}(s_k)| \leq \mathcal{\tilde{O}}\left(\dfrac{\alpha G C t_{\mathrm{mix}}}{t_{\mathrm{hit}}}T^{\frac{1}{4}}\right)
    \end{align*}
    where $\bar{\theta}_k$ denotes some convex combination of $\theta_k$ and $\theta_{k+1}$, $C\triangleq \sqrt{A}G t_{\mathrm{mix}}+\sqrt{Lt_{\mathrm{mix}}t_{\mathrm{hit}}}$, and $\{s_k\}_{k=1}^K$ is an arbitrary sequence of states.
\end{lemma}

Lemma \ref{lemma_last} can be interpreted as the stability results of our algorithm. It essentially states that the policy parameters are updated such that the average difference between consecutive average reward and value functions decreases with the horizon, $T$. Using the above result, we now prove our regret guarantee. 

\begin{theorem}
    \label{thm_regret}
    If assumptions \ref{ass_1}, \ref{ass_score}, \ref{ass_transfer_error}, and \ref{ass_4} hold, $J(\cdot)$ is $L$-smooth, and $T$ is sufficiently large, then our proposed Algorithm \ref{alg:PG_MAG} achieves the following expected regret bound with learning rate $\alpha=\frac{1}{4L}$.
    \begin{align}
        \label{eq_regret}
        \begin{split}
            &\mathbf{E}\left[\mathrm{Reg}_{T}\right] \leq T\sqrt{\epsilon_{\mathrm{bias}}} + \mathcal{O}(t_{\mathrm{mix}})+ \mathcal{\tilde{O}}\left(\dfrac{BC^2}{L}\sqrt{T}\right)\\
            &+\mathcal{\tilde{O}}\left(GC\left(1+\dfrac{1}{\mu_F}\right)T^{\frac{3}{4}}\right)+ \mathcal{\tilde{O}}\left(\dfrac{ G C t_{\mathrm{mix}}}{Lt_{\mathrm{hit}}}T^{\frac{1}{4}}\right)\\
            &+\mathcal{\tilde{O}}\left(Lt_{\mathrm{mix}}t_{\mathrm{hit}}\mathbf{E}_{s\sim d^{\pi^*}}[KL(\pi^*(\cdot\vert s)\Vert\pi_{\theta_1}(\cdot\vert s))]\sqrt{T}\right)
        \end{split}
    \end{align}
    where the term $C$ is the same as defined in Lemma \ref{lemma_last}.
\end{theorem}

Theorem \ref{thm_regret} shows that the expected regret of Algorithm \ref{alg:PG_MAG} is bounded by $\tilde{\mathcal{O}}(T^{\frac{3}{4}}+T\sqrt{\epsilon_{\mathrm{bias}}})$. It also shows how other parameters such as  $t_{\mathrm{mix}}$, and  $t_{\mathrm{hit}}$ influence the regret value. Note that the term related to $\epsilon_{\mathrm{bias}}$ is unavoidable in our setting due to the incompleteness of the general parameterized policy set.


\section{Conclusion}

In this paper, we proposed an algorithm based on the vanilla policy gradient for reinforcement learning in an infinite horizon average reward setting. Unlike the recent works on this setting which require the MDP to be tabular or have a linear structure, we assume the framework of general parametrization of the policy. We show that the proposed algorithm converges to the neighborhood of the global optimum with rate $\mathcal{O}(T^{-1/4})$, which matches the result of vanilla policy gradient with general parametrization in discounted reward setting. We use this convergence result to further show that our algorithm achieves a regret of $\tilde{\mathcal{O}}(T^{\frac{3}{4}})$. 

We note that this paper unveils numerous promising directions for future research. These avenues encompass exploring the possibility of relaxing the assumption of ergodic MDPs to weakly communicating MDPs, refining regret bounds for enhanced performance, deriving more robust lower bounds for the general parametrization, and extending the problem domain to incorporate constraints.

\section{Acknowledgements}
This work was supported in part by the U.S. National Science Foundation under Grant CCF-2149588 and Cisco Inc.


\bibliography{ref.bib}


\onecolumn

\input{appendix}

\end{document}

%% file: appendix.tex

\section{Proofs for the Section of Global Convergence Analysis}


\subsection{Proof of Lemma \ref{lemma_grad_compute}}

\begin{proof}
    Using (\ref{eq_V_Q}), we arrive at the following.
    \begin{equation}
        \label{eq_22}
        \begin{aligned}
            &\nabla_{\theta} V^{\pi_{\theta_k}}(s) =\nabla_{\theta}\bigg(\sum_{a}\pi_{\theta}(a|s)Q^{\pi_{\theta}}(s, a)\bigg)\\
            &=\sum_{a}\bigg(\nabla_{\theta}\pi_{\theta}(a|s)\bigg)Q^{\pi_{\theta}}(s, a)+\sum_{a}\pi_\theta(a|s)\nabla_{\theta} Q^{\pi_{\theta}}(s, a)\\
            &\overset{(a)}=\sum_{a}\pi_{\theta}(a|s)\bigg(\nabla_{\theta}\log\pi_{\theta}(a|s)\bigg)Q^{\pi_{\theta}}(s, a)+\sum_{a}\pi_\theta(a|s)\nabla_{\theta} \bigg(r(s, a)-J(\theta)+\sum_{s'}P(s'|s, a)V^{\pi_{\theta}}(s')\bigg)\\
            &=\sum_{a}\pi_\theta(a|s)\bigg(\nabla_{\theta}\log\pi_{\theta}(a|s)\bigg)Q^{\pi_{\theta}}(s, a)+\sum_{a}\pi_\theta(a|s) \bigg(\sum_{s'}P(s'|s,a)\nabla_{\theta} V^{\pi_{\theta}}(s')\bigg)-\nabla_{\theta}J(\theta)
        \end{aligned}
    \end{equation}
    where the step (a) is a consequence of $\nabla_{\theta}\log\pi_{\theta}=\frac{\nabla\pi_{\theta}}{\pi_{\theta}}$ and the Bellman equation. Multiplying both sides by $d^{\pi_{\theta}}(s)$, taking a sum over $s\in\mathcal{S}$, and rearranging the terms, we obtain the following.
    \begin{align}
        \begin{split}
            \nabla_{\theta}J(\theta)&=\sum_{s}d^{\pi_{\theta}}(s)\nabla_{\theta}J(\theta)\\
            &=\sum_{s}d^{\pi_{\theta}}(s)\sum_{a}\pi_\theta(a|s)\bigg(\nabla_{\theta}\log\pi_{\theta}(a|s)\bigg)Q^{\pi_{\theta}}(s, a)\\
            &+\sum_{s}d^{\pi_{\theta}}(s)\sum_{a}\pi_\theta(a|s) \bigg(\sum_{s'}P(s'|s,a)\nabla_{\theta} V^{\pi_{\theta}}(s')\bigg) - \sum_{s}d^{\pi_{\theta}}(s)\nabla_{\theta}V^{\pi_\theta}(s)\\
            &\overset{}{=}\mathbf{E}_{s\sim d^{\pi_\theta}, a\sim \pi_\theta(\cdot|s)}\bigg[Q^{\pi_{\theta}}(s, a)\nabla_{\theta}\log\pi_{\theta}(a|s)\bigg]+\sum_{s}d^{\pi_{\theta}}(s) \sum_{s'}P^{\pi_\theta}(s'|s)\nabla_{\theta} V^{\pi_{\theta}}(s') - \sum_{s}d^{\pi_{\theta}}(s)\nabla_{\theta}V^{\pi_\theta}(s)\\
            &\overset{(a)}{=}\mathbf{E}_{s\sim d^{\pi_\theta}, a\sim \pi_\theta(\cdot|s)}\bigg[Q^{\pi_{\theta}}(s, a)\nabla_{\theta}\log\pi_{\theta}(a|s)\bigg]+ \sum_{s'}d^{\pi_{\theta}}(s')\nabla_{\theta} V^{\pi_{\theta}}(s') - \sum_{s}d^{\pi_{\theta}}(s)\nabla_{\theta}V^{\pi_\theta}(s)\\
            &=\mathbf{E}_{s\sim d^{\pi_\theta}, a\sim \pi_\theta(\cdot|s)}\bigg[Q^{\pi_{\theta}}(s, a)\nabla_{\theta}\log\pi_{\theta}(a|s)\bigg]
        \end{split}
    \end{align}
    where $(a)$ uses the fact that $d^{\pi_\theta}$ is a stationary distribution. Note that,
    \begin{equation}
        \begin{aligned}
            \mathbf{E}_{s\sim d^{\pi_{\theta}},a\sim\pi_{\theta}(\cdot|s)}&\bigg[ V^{\pi_{\theta}}(s)\nabla\log\pi_{\theta}(a|s)\bigg]\\
            &=\mathbf{E}_{s\sim d^{\pi_{\theta}}}\left[ \sum_{a\in\mathcal{A}}V^{\pi_{\theta}}(s)\nabla_{\theta}\pi_{\theta}(a|s)\right]\\
            &=\mathbf{E}_{s\sim d^{\pi_{\theta}}}\left[ \sum_{a\in\mathcal{A}}V^{\pi_{\theta}}(s)\nabla_{\theta}\pi_{\theta}(a|s)\right]\\
            &=\mathbf{E}_{s\sim d^{\pi_{\theta}}}\bigg[ V^{\pi_{\theta}}(s)\nabla_{\theta}\left(\sum_{a\in\mathcal{A}}\pi_{\theta}(a|s)\right)\bigg] = \mathbf{E}_{s\sim d^{\pi_{\theta}}}\bigg[ V^{\pi_{\theta}}(s)\nabla_{\theta}(1)\bigg] = 0
        \end{aligned}
    \end{equation}
    We can, therefore, replace the function $Q^{\pi_{\theta}}$ in the policy gradient with the advantage function $A^{\pi_{\theta}}(s, a)=Q^{\pi_{\theta}}(s, a)-V^{\pi_{\theta}}(s)$, $\forall (s, a)\in\mathcal{S}\times \mathcal{A}$. Thus,
    \begin{equation}
        \nabla_{\theta} J(\theta)=\mathbf{E}_{s\sim d^{\pi_{\theta}},a\sim\pi_{\theta}(\cdot|s)}\bigg[ A^{\pi_{\theta}}(s,a)\nabla_{\theta}\log\pi_{\theta}(a|s)\bigg]
    \end{equation}    
\end{proof}


\subsection{Proof of Lemma \ref{lemma_good_estimator}}

\begin{proof}
    The proof runs along the line of \citep[Lemma 6]{wei2020model}. Consider the $k$th epoch and assume that $\pi_{\theta_k}$ is denoted as $\pi$ for notational convenience. Let, $M$ be the number of disjoint sub-trajectories of length $N$ that start with the state $s$ and are at least $N$ distance apart (found by Algorithm \ref{alg:estQ}). Let, $y_{k, i}$ be the sum of observed rewards of the $i$th sub-trajectory and $\tau_i$ be its starting time. The advantage function estimate is,
    \begin{align}
        \label{def_A_hat_appndx}
        \hat{A}^{\pi}(s, a) = \begin{cases}
            \dfrac{1}{\pi(a|s)}\left[\dfrac{1}{M}\sum_{i=1}^M y_{k,i}\mathrm{1}(a_{\tau_i}=a)\right] - \dfrac{1}{M}\sum_{i=1}^M y_{k,i}~&\text{if}~M>0\\
            0~&\text{if}~M=0
        \end{cases}
    \end{align}

    Note the following,
    \begin{align}
        \begin{split}
           &\mathbf{E}\left[y_{k,i}\bigg|s_{\tau_i}=s, a_{\tau_i}=a\right]\\
           &=r(s, a) + \mathbf{E}\left[\sum_{t=\tau_i+1}^{\tau_i+N}r(s_t, a_t)\bigg| s_{\tau_i}=s, a_{\tau_i}=a\right]\\
           &=r(s, a) + \sum_{s'}P(s'|s, a)\mathbf{E}\left[\sum_{t=\tau_i+1}^{\tau_i+N}r(s_t, a_t)\bigg| s_{\tau_i+1}=s'\right]\\
           &=r(s, a) + \sum_{s'}P(s'|s, a)\left[\sum_{j=0}^{N-1}(P^{\pi})^j(s', \cdot)\right]^Tr^{\pi}\\
           &=r(s, a) + \sum_{s'}P(s'|s, a)\left[\sum_{j=0}^{N-1}(P^{\pi})^j(s', \cdot)-d^{\pi}\right]^Tr^{\pi} + N(d^{\pi})^Tr^{\pi}\\
           &\overset{(a)}{=}r(s, a) + \sum_{s'}P(s'|s, a)\left[\sum_{j=0}^{\infty}(P^{\pi})^j(s', \cdot)-d^{\pi}\right]^Tr^{\pi} + NJ^{\pi}-\underbrace{\sum_{s'}P(s'|s, a)\left[\sum_{j=N}^{\infty}(P^{\pi})^j(s', \cdot)-d^{\pi}\right]^Tr^{\pi}}_{\triangleq \mathrm{E}^{\pi}_T(s, a)}\\
           &\overset{(b)}{=} r(s, a) + \sum_{s'}P(s'|s, a)V^{\pi}(s') + NJ^{\pi}-\mathrm{E}^{\pi}_T(s, a)\\
           &\overset{(c)}{=} Q^{\pi}(s, a) + (N+1)J^{\pi} - \mathrm{E}^{\pi}_T(s, a)
        \end{split}
    \end{align}
    where $(a)$ follows from the definition of $J^{\pi}$ as given in $(\ref{eq_r_pi_theta})$, $(b)$ is an application of the definition of $V^{\pi}$ given in $(\ref{def_v_pi_theta_s})$, and $(c)$ follows from the Bellman equation. Define the following quantity.
    \begin{align}
        \label{def_error_1}
        \delta^{\pi}(s, T) \triangleq \sum_{t=N}^{\infty}\norm{(P^{\pi})^t({s,\cdot}) - d^{\pi}}_1 ~~\text{where} ~N=4t_{\mathrm{mix}}(\log_2 T)
    \end{align}
    
    Using Lemma \ref{lemma_aux_3}, we get $\delta^{\pi}(s, T)\leq \frac{1}{T^3}$ which implies, $|\mathrm{E}^{\pi}_T(s, a)|\leq \frac{1}{T^3}$. Observe that,
    \begin{align}
        \label{eq_appndx_47}
        \begin{split}
            &\mathbf{E}\left[\left(\dfrac{1}{\pi(a|s)}y_{k, i}\mathrm{1}(a_{\tau_i}=a) - y_{k, i}\right)\bigg| s_{\tau_i}=s\right] \\
            &= \mathbf{E}\left[y_{k, i}\bigg| s_{\tau_i}=s, a_{\tau_i}=a\right] - \sum_{a'}\pi(a'|s)\mathbf{E}\left[y_{k, i}\bigg| s_{\tau_i}=s, a_{\tau_i}=a^{'}\right]\\
            &=Q^{\pi}(s, a) + (N+1)J^{\pi} - \mathrm{E}^{\pi}_T(s, a) - \sum_{a'}\pi(a'|s)[Q^{\pi}(s, a) + (N+1)J^{\pi} - \mathrm{E}^{\pi}_T(s, a)]\\
            &=Q^{\pi}(s, a)-V^{\pi}(s)-\left[\mathrm{E}_T(s, a) - \sum_{a'}\pi(a'|s)\mathrm{E}_T^{\pi}(s, a')\right]\\
            &= A^{\pi}(s, a) -\Delta^{\pi}_T(s, a)
        \end{split}
    \end{align}
    where $\Delta^{\pi}_T(s, a)\triangleq\mathrm{E}_T(s, a) - \sum_{a'}\pi(a'|s)\mathrm{E}_T^{\pi}(s, a')$. Using the bound on $\mathrm{E}^{\pi}_T(s, a)$, one can show that, $|\Delta_T^{\pi}(s, a)|\leq \frac{2}{T^3}$, which implies,
    \begin{align}
        \label{eq_appndx_48}
        \left|\mathbf{E}\left[\left(\dfrac{1}{\pi(a|s)}y_{k, i}\mathrm{1}(a_{\tau_i}=a) - y_{k, i}\right)\bigg| s_{\tau_i}=s\right] - A^{\pi}(s, a)\right|\leq |\Delta_T^{\pi}(s, a)|\leq\dfrac{2}{T^3}
    \end{align}

    Note that \eqref{eq_appndx_48} cannot be directly used to bound the bias of $\hat{A}^{\pi}(s, a)$. This is because the random variable $M$ is correlated with the reward variables $\{y_{k, i}\}_{i=1}^M$. To decorrelate these two random variables, imagine an MDP where the state distribution rejuvenates to the stationary distribution, $d^{\pi}$ after exactly $N$ time steps since the completion of a sub-trajectory. In other words, if a sub-trajectory starts at $\tau_{i}$, and ends at $\tau_i+N$, then the system `rests' for additional $N$ steps before rejuvenating with the state distribution, $d^{\pi}$ at $\tau_i+2N$. Clearly, the wait time between the rejuvenation after the $(i-1)$th sub-trajectory and the start of the $i$th sub-trajectory is, $w_{i}=\tau_{i}-(\tau_{i-1}+2N)$, $i>1$. Let $w_1$ be the time between the start time of the $k$th epoch and the start time of the first sub-trajectory. Note that,

    $(a)$ $w_1$ only depends on the initial state, $s_{(k-1)H}$ and the induced transition function, $P^{\pi}$,

    $(b)$ $w_i$, where $i>1$, depends on the stationary distribution, $d^{\pi}$, and the induced transition function, $P^{\pi}$,

    $(c)$ $M$ only depends on $\{w_1, w_2, \cdots\}$ as other segments of the epoch have fixed length, $2N$.

    Clearly, in this imaginary MDP, the sequence, $\{w_1, w_2, \cdots\}$, and hence, $M$ is independent of $\{y_{k,1}, y_{k, 2}, \cdots\}$. Let, $\mathbf{E}'$ denote the expectation operation and $\mathrm{Pr}'$ denote the probability of events in this imaginary system. Define the following.
    \begin{align}
    \label{def_delta_i}
        \Delta_i \triangleq \dfrac{y_{k,i}\mathrm{1}(a_{\tau_i}=a)}{\pi(a|s)} - y_{k, i} - A^{\pi}(s, a) + \Delta^{\pi}_T(s, a)
    \end{align}
    where $\Delta^{\pi}_T(s, a)$ is defined in $(\ref{eq_appndx_47})$. Note that we have suppressed the dependence on $T$, $s, a$, and $\pi$ while defining $\Delta_i$ to remove clutter. Using $(\ref{eq_appndx_47})$, one can write $ \mathbf{E}'\left[\Delta_i(s, a)|\{w_i\}\right]=0$. Moreover, 
    \begin{align}
        \label{eq_appndx_50}
       \begin{split}
           &\mathbf{E}'\left[\left(\hat{A}^{\pi}(s, a) - A^{\pi}(s, a)\right)^2\right]\\ 
           &= \mathbf{E}'\left[\left(\hat{A}^{\pi}(s, a) - A^{\pi}(s, a)\right)^2\bigg| M>0\right]\times \mathrm{Pr}'(M>0) + \left(A^{\pi}(s, a)\right)^2\times \mathrm{Pr}'(M=0)\\
           &= \mathbf{E}'\left[\left(\dfrac{1}{M}\sum_{i=1}^M\Delta_i - \Delta_T^{\pi}(s, a)\right)^2\bigg| M>0\right]\times \mathrm{Pr}'(M>0) + \left(A^{\pi}(s, a)\right)^2\times \mathrm{Pr}'(M=0)\\
           & \overset{}{\leq} 2\mathbf{E}_{\{w_i\}}'\left[\mathbf{E}'\left[\left(\dfrac{1}{M}\sum_{i=1}^M\Delta_i \right)^2\bigg| \{w_i\}\right]\bigg| w_1\leq H-N\right]\times \mathrm{Pr}'(w_1\leq H-N) + 2\left(\Delta_T^{\pi}(s, a)\right)^2+\left(A^{\pi}(s, a)\right)^2\times \mathrm{Pr}'(M=0)\\
           & \overset{(a)}{\leq} 2\mathbf{E}_{\{w_i\}}'\left[\dfrac{1}{M^2}\sum_{i=1}^M \mathbf{E}'\left[\Delta_i^2\big|\{w_i\}\right]\bigg| w_1\leq H-N\right]\times \mathrm{Pr}'(w_1\leq H-N) + \dfrac{8}{T^6} +\left(A^{\pi}(s, a)\right)^2\times \mathrm{Pr}'(M=0)\\
       \end{split}
    \end{align}
    where $(a)$ uses the bound $|\Delta_T^{\pi}(s, a)|\leq \frac{2}{T^3}$ derived in $(\ref{eq_appndx_48})$, and the fact that $\{\Delta_i\}$ are zero mean independent random variables conditioned on $\{w_i\}$. Note that $|y_{k, i}|\leq N$ almost surely, $|A^{\pi}(s, a)|\leq \mathcal{O}(t_{\mathrm{mix}})$ via Lemma \ref{lemma_aux_2}, and $|\Delta^{\pi}_T(s, a)|\leq \frac{2}{T^3}$ as shown in $(\ref{eq_appndx_48})$. Combining, we get, $\mathbf{E}'[|\Delta_i|^2\big|\{w_i\}]\leq \mathcal{O}(N^2/\pi(a|s))$ (see the definition of $\Delta_i$ in (\ref{def_delta_i})). Invoking this bound into $(\ref{eq_appndx_50})$, we get the following result.
    \begin{align}
        \label{eq_appndx_51_}
        \begin{split}
        \mathbf{E}'&\left[\left(\hat{A}^{\pi}(s, a) - A^{\pi}(s, a)\right)^2\right]\leq 2\mathbf{E}'\left[\dfrac{1}{M}\bigg|w_1\leq H-N\right]\mathcal{O}\left(\dfrac{N^2}{\pi(a|s)}\right)+\dfrac{8}{T^6}+\mathcal{O}(t_{\mathrm{mix}}^2)\times \mathrm{Pr}'(w_1>H-N)\\
        \end{split}
    \end{align}

    Note that, one can use Lemma \ref{lemma_aux_4} to bound the following violation probability.
    \begin{align}
        \label{eq_appndx_52_}
        \mathrm{Pr}'(w_1>H-N)\leq \left(1-\dfrac{3d^{\pi}(s)}{4}\right)^{4t_{\mathrm{hit}}\sqrt{T}(\log T)-1}\overset{(a)}{\leq} \left(1-\dfrac{3d^{\pi}(s)}{4}\right)^{\dfrac{4}{d^{\pi}(s)}(\log T)}\leq \dfrac{1}{T^3}
    \end{align}
    where $(a)$ follows from the fact that $4t_{\mathrm{hit}}\sqrt{T}(\log_2 T) - 1 \geq \frac{4}{d^{\pi}(s)}\log_2 T$ for sufficiently large $T$. Finally, note that, if $M<M_0$, where $M_0$ is defined as,
    \begin{align}
        M_0\triangleq \dfrac{H-N}{2N+ \dfrac{4N\log T}{d^{\pi}(s)}}
    \end{align}
    then there exists at least one $w_i$ that is larger than $4N\log_2 T/d^{\pi}(s)$ which can happen with the following maximum probability according to Lemma \ref{lemma_aux_4}.
    \begin{align}
        \mathrm{Pr}'\left(M<M_0\right) \leq \left(1-\dfrac{3d^{\pi}(s)}{4}\right)^{\frac{4\log T}{d^{\pi(s)}}}\leq \dfrac{1}{T^3}
    \end{align}

    The above probability bound can be used to obtain the following result,
    \begin{align}
        \label{eq_appndx_55_}
        \begin{split}
            \mathbf{E}'\left[\dfrac{1}{M}\bigg| M>0\right]=\dfrac{\sum_{m=1}^{\infty}\dfrac{1}{m}\mathrm{Pr}'(M=m)}{\mathrm{Pr}'(M>0)}&\leq \dfrac{1\times \mathrm{Pr}'(M\leq M_0)+\dfrac{1}{M_0}\mathrm{Pr}'(M>M_0)}{\mathrm{Pr}'(M>0)}\\
            &\leq  \dfrac{\dfrac{1}{T^3}+\dfrac{2N+\dfrac{4N \log T}{d^{\pi}(s)}}{H-N}}{1-\dfrac{1}{T^3}}\leq \mathcal{O}\left(\dfrac{N\log T}{H d^{\pi}(s)}\right)
        \end{split}
    \end{align}

    Injecting $(\ref{eq_appndx_52_})$ and $(\ref{eq_appndx_55_})$ into $(\ref{eq_appndx_51_})$, we finally obtain the following.
    \begin{align}
        \label{eq_appndx_56_}
        \mathbf{E}'&\left[\left(\hat{A}^{\pi}(s, a) - A^{\pi}(s, a)\right)^2\right]\leq \mathcal{O}\left(\dfrac{N^3\log T}{H d^{\pi}(s)\pi(a|s)}\right)=\mathcal{O}\left(\dfrac{N^3t_{\mathrm{hit}}\log T}{H \pi(a|s)}\right)=\mathcal{O}\left(\dfrac{t^2_{\mathrm{mix}}(\log T)^2}{\sqrt{T}\pi(a|s)}\right)
    \end{align}

    Eq. $(\ref{eq_appndx_56_})$ shows that our desired inequality is satisfied in the imaginary system. We now need a mechanism to translate this result to our actual MDP. Notice that, we can write $(\hat{A}^{\pi}(s, a)-A^{\pi}(s, a))^2=f(X)$ where $X=(M, \tau_1, \mathcal{T}_1, \cdots, \tau_M, \mathcal{T}_M)$, and $\mathcal{T}_i = (a_{\tau_i}, s_{\tau_i+1}, a_{\tau_i+1}, \cdots, s_{\tau_i+N}, a_{\tau_i+N})$. We have,
    \begin{align}
        \label{eq_appndx_57_}
        \dfrac{\mathbf{E}[f(X)]}{\mathbf{E}'[f(X)]} = \dfrac{\sum_{X} f(X)\mathrm{Pr}(X)}{\sum_{X} f(X)\mathrm{Pr}'(X)}\leq \max_{X}\dfrac{\mathrm{Pr}(X)}{\mathrm{Pr'}(X)}
    \end{align}

    The last inequality uses the non-negativity of $f(\cdot)$. Observe that, for a fixed sequence, $X$, we have,
    \begin{align}
        \begin{split}
            \mathrm{Pr}(X) = &\mathrm{Pr}(\tau_1)\times \mathrm{Pr}(\mathcal{T}_1|\tau_1)\times \mathrm{Pr}(\tau_2|\tau_1, \mathcal{T}_1)\times \mathrm{Pr}(\mathcal{T}_2|\tau_2) \times \cdots \\
            &\times \mathrm{Pr}(\tau_M|\tau_{M-1}, \mathcal{T}_{M-1})\times \mathrm{Pr}(\mathcal{T}_M|\tau_M)\times \mathrm{Pr}(s_t\neq s, \forall t\in[\tau_M+2N, kH-N]|\tau_M, \mathcal{T}_M),
        \end{split}\\
        \begin{split}
            \mathrm{Pr}'(X) = &\mathrm{Pr}(\tau_1)\times \mathrm{Pr}(\mathcal{T}_1|\tau_1)\times \mathrm{Pr}'(\tau_2|\tau_1, \mathcal{T}_1)\times \mathrm{Pr}(\mathcal{T}_2|\tau_2) \times \cdots \\
            &\times \mathrm{Pr}'(\tau_M|\tau_{M-1}, \mathcal{T}_{M-1})\times \mathrm{Pr}(\mathcal{T}_M|\tau_M)\times \mathrm{Pr}(s_t\neq s, \forall t\in[\tau_M+2N, kH-N]|\tau_M, \mathcal{T}_M),
        \end{split}
    \end{align}

    Thus, the difference between $\mathrm{Pr}(X)$ and $\mathrm{Pr}'(X)$ arises because $\mathrm{Pr}(\tau_{i+1}|\tau_i, \mathcal{T}_i)\neq \mathrm{Pr}'(\tau_{i+1}|\tau_i, \mathcal{T}_i)$, $\forall i\in\{1, \cdots, M-1\}$. Note that the ratio of these two terms can be bounded as follows,
    \begin{align}
        \begin{split}
            \dfrac{\mathrm{Pr}(\tau_{i+1}|\tau_i, \mathcal{T}_i)}{\mathrm{Pr}'(\tau_{i+1}|\tau_i, \mathcal{T}_i)}&=\dfrac{\sum_{s'\neq s} \mathrm{Pr}(s_{\tau_i+2N}=s'|\tau_i, \mathcal{T}_i)\times \mathrm{Pr}(s_t\neq s, \forall t\in [\tau_i+2N, \tau_{i+1}-1], s_{\tau_{i+1}}=s|s_{\tau_i+2N}=s')}{\sum_{s'\neq s} \mathrm{Pr}'(s_{\tau_i+2N}=s'|\tau_i, \mathcal{T}_i)\times \mathrm{Pr}(s_t\neq s, \forall t\in [\tau_i+2N, \tau_{i+1}-1], s_{\tau_{i+1}}=s|s_{\tau_i+2N}=s')}\\
            &\leq \max_{s'}\dfrac{\mathrm{Pr}(s_{\tau_i+2N}=s'|\tau_i, \mathcal{T}_i)}{\mathrm{Pr}'(s_{\tau_i+2N}=s'|\tau_i, \mathcal{T}_i)}\\
            &=\max_{s'}1+\dfrac{\mathrm{Pr}(s_{\tau_i+2N}=s'|\tau_i, \mathcal{T}_i)-d^{\pi}(s')}{d^{\pi}(s')}\overset{(a)}{\leq} \max_{s'}1+\dfrac{1}{T^3d^{\pi}(s')}\leq 1+\dfrac{t_{\mathrm{hit}}}{T^3}\leq 1+\dfrac{1}{T^2}
        \end{split}
    \end{align}
    where $(a)$ is a consequence of Lemma \ref{lemma_aux_3}. We have,
    \begin{align}
        \label{eq_appndx_61_}
        \dfrac{\mathrm{Pr}(X)}{\mathrm{Pr}'(X)}\leq \left(1+\dfrac{1}{T^2}\right)^M\leq e^{\frac{M}{T^2}}\overset{(a)}{\leq} e^{\frac{1}{T}}\leq \mathcal{O}\left(1+\dfrac{1}{T}\right) 
    \end{align}
    where $(a)$ uses the fact that $M\leq T$. Combining $(\ref{eq_appndx_57_})$ and $(\ref{eq_appndx_61_})$, we get,
    \begin{align}
        \begin{split}
            \mathbf{E}\left[\left(\hat{A}^{\pi}(s, a) - A^{\pi}(s, a)\right)^2\right]&\leq \mathcal{O}\left(1+\dfrac{1}{T}\right)\mathbf{E}'\left[\left(\hat{A}^{\pi}(s, a) - A^{\pi}(s, a)\right)^2\right]\\
            &\overset{(a)}{\leq} \mathcal{O}\left(\dfrac{t^2_{\mathrm{mix}}(\log T)^2}{\sqrt{T}\pi(a|s)}\right)
        \end{split}
    \end{align}
    where $(a)$ follows from $(\ref{eq_appndx_56_})$. This concludes the lemma.
\end{proof}


\subsection{Proof of Lemma \ref{lemma_grad_est_bias}}

\begin{proof} 
    Recall from Eq. \eqref{eq_grad_estimate} that,
    \begin{align}
        &\omega_k= \dfrac{1}{H}\sum_{t=t_k}^{t_{k+1}-1}\hat{A}^{\pi_{\theta_k}}(s_{t}, a_{t})\nabla_{\theta}\log \pi_{\theta_k}(a_{t}|s_{t}),
    \end{align}
    
    Define the following quantity,
    \begin{align}
        &\bar{\omega}_k= \dfrac{1}{H}\sum_{t=t_k}^{t_{k+1}-1}A^{\pi_{\theta_k}}(s_{t}, a_{t})\nabla_{\theta}\log \pi_{\theta_k}(a_{t}|s_{t})
    \end{align}
    where $t_k=(k-1)H$ is the starting time of the $k$th epoch. Note that the true gradient is given by, 
    \begin{align}
        \nabla_{\theta}J(\theta_k)=\mathbf{E}_{s\sim d^{\pi_{\theta_k}}, a\sim\pi_{\theta_k}(\cdot|s)}\left[A^{\pi_{\theta_k}}(s, a)\nabla_{\theta}\log\pi_{\theta}(a|s)\right]
    \end{align}
    
    Moreover, using Assumption \ref{ass_score} and Lemma \ref{lemma_aux_3}, one can prove that $|A^{\pi_{\theta_k}}(s, a)\nabla_{\theta}\log\pi_{\theta}(a|s)|\leq \mathcal{O}(t_{\mathrm{mix}}G)$, $\forall (s, a)\in \mathcal{S}\times \mathcal{A}$ which implies    $|\nabla_{\theta}J(\theta_k)|\leq \mathcal{O}(t_{\mathrm{mix}}G)$. Applying Lemma \ref{lemma_aux_6}, we, therefore, arrive at,
    \begin{align}
        \label{eq_appndx_67_}
        \mathbf{E}\left[\norm{\bar{\omega}_k-\nabla_{\theta}J(\theta_k)}^2\right]\leq \mathcal{O}\left(G^{2}t^2_{\mathrm{mix}}\log T\right)\times \mathcal{O}\left(\dfrac{t_{\mathrm{mix}}\log T}{H}\right)=\mathcal{O}\left(\dfrac{G^2t_{\mathrm{mix}}^2}{t_{\mathrm{hit}}\sqrt{T}}\right)
    \end{align}
    
    We would like to point out that Lemma \ref{lemma_aux_6} was also used in \citep{suttle2023beyond} to analyze their actor-critic algorithm. Finally, the difference, $\mathbf{E}\norm{\omega_k-\bar{\omega}_k}^2$ can be bounded as follows.
    \begin{align}
        \label{eq_appndx_68_}
        \begin{split}
            \mathbf{E}\norm{\omega_k-\bar{\omega}_k}^2&=\mathbf{E}\left[\norm{\dfrac{1}{H}\sum_{t=t_k}^{t_{k+1}-1}\hat{A}^{\pi_{\theta_k}}(s_{t}, a_{t})\nabla_{\theta}\log \pi_{\theta_k}(a_{t}|s_{t})-\dfrac{1}{H}\sum_{t=t_k}^{t_{k+1}-1}\hat{A}^{\pi_{\theta_k}}(s_{t}, a_{t})\nabla_{\theta}\log \pi_{\theta_k}(a_{t}|s_{t})}^2\right]\\
            &\overset{(a)}{\leq} \dfrac{G^2}{H}\sum_{t=t_k}^{t_{k+1}-1}\mathbf{E}\left[\left(\hat{A}^{\pi_{\theta_k}}(s_t, a_t)-A^{\pi_{\theta_k}}(s_t, a_t)\right)^2\right]\\
            &\leq \dfrac{G^2}{H}\sum_{t=t_k}^{t_{k+1}-1}\mathbf{E}\left[\sum_{a}\pi_{\theta_k}(a|s_t)\mathbf{E}\left[\left(\hat{A}^{\pi_{\theta_k}}(s_t, a)-A^{\pi_{\theta_k}}(s_t, a)\right)^2\bigg| s_t\right]\right]\\
            &\overset{(b)}{\leq}\mathcal{O}\left(\dfrac{AG^2t^2_{\mathrm{mix}}(\log T)^2}{\sqrt{T}}\right)
        \end{split}
    \end{align}
    where $(a)$ follows from Assumption \ref{ass_score} and Jensen's inequality whereas $(b)$ follows from Lemma \ref{lemma_good_estimator}. Combining, $(\ref{eq_appndx_67_})$ and $(\ref{eq_appndx_68_})$, we conclude the result.
\end{proof}

    
\subsection{Proof of Lemma \ref{lem_performance_diff}}

\begin{proof}
    Using the Lemma \ref{lemma_aux_5}, it is obvious to see that
    \begin{equation}
        \begin{aligned}
            J^{\pi}-J^{\pi'}&=\sum_{s}\sum_{a}d^{\pi}(s)(\pi(a|s)-\pi'(a|s))Q^{\pi'}(s,a)\\
            &=\sum_{s}\sum_{a}d^{\pi}(s)\pi(a|s)Q^{\pi'}(s,a)-\sum_{s}d^{\pi}(s)V^{\pi'}(s)\\
            &\overset{(a)}{=}\sum_{s}\sum_{a}d^{\pi}(s)\pi(a|s)Q^{\pi'}(s,a)-\sum_{s}\sum_{a}d^{\pi}(s)\pi(a|s)V^{\pi'}(s)\\
            &=\sum_{s}\sum_{a}d^{\pi}(s)\pi(a|s)[Q^{\pi'}(s,a)-V^{\pi'}(s)]\\
            &=\mathbf{E}_{s\sim d^{\pi}}\mathbf{E}_{a\sim\pi(\cdot\vert s)}\big[A^{\pi'}(s, a)\big]
        \end{aligned}
    \end{equation}
    Equation (a) follows from the fact that $\sum_{a}\pi(a|s)=1$, $\forall s\in\mathcal{S}$.
\end{proof}


\subsection{Proof of Lemma \ref{lem_framework}}

\begin{proof}
    We start with the definition of KL divergence.
    \begin{equation}
	\begin{aligned}
            &\mathbf{E}_{s\sim d^{\pi^*}}[KL(\pi^*(\cdot\vert s)\Vert\pi_{\theta_k}(\cdot\vert s))-KL(\pi^*(\cdot\vert s)\Vert\pi_{\theta_{k+1}}(\cdot\vert s))]\\
            &=\mathbf{E}_{s\sim d^{\pi^*}}\mathbf{E}_{a\sim\pi^*(\cdot\vert s)}\bigg[\log\frac{\pi_{\theta_{k+1}(a\vert s)}}{\pi_{\theta_k}(a\vert s)}\bigg]\\
            &\overset{(a)}\geq\mathbf{E}_{s\sim d^{\pi^*}}\mathbf{E}_{a\sim\pi^*(\cdot\vert s)}[\nabla_\theta\log\pi_{\theta_k}(a\vert s)\cdot(\theta_{k+1}-\theta_k)]-\frac{B}{2}\Vert\theta_{k+1}-\theta_k\Vert^2\\
            &=\alpha\mathbf{E}_{s\sim d^{\pi^*}}\mathbf{E}_{a\sim\pi^*(\cdot\vert s)}[\nabla_{\theta}\log\pi_{\theta_k}(a\vert s)\cdot\omega_k]-\frac{B\alpha^2}{2}\Vert\omega_k\Vert^2\\
            &=\alpha\mathbf{E}_{s\sim d^{\pi^*}}\mathbf{E}_{a\sim\pi^*(\cdot\vert s)}[\nabla_\theta\log\pi_{\theta_k}(a\vert s)\cdot\omega^*_k]+\alpha\mathbf{E}_{s\sim d^{\pi^*}}\mathbf{E}_{a\sim\pi^*(\cdot\vert s)}[\nabla_\theta\log\pi_{\theta_k}(a\vert s)\cdot(\omega_k-\omega^*_k)]-\frac{B\alpha^2}{2}\Vert\omega_k\Vert^2\\
            &=\alpha[J^{*}-J(\theta_k)]+\alpha\mathbf{E}_{s\sim d^{\pi^*}}\mathbf{E}_{a\sim\pi^*(\cdot\vert s)}[\nabla_\theta\log\pi_{\theta_k}(a\vert s)\cdot\omega^*_k]-\alpha[J^{*}-J(\theta_k)]\\
            &+\alpha\mathbf{E}_{s\sim d^{\pi^*}}\mathbf{E}_{a\sim\pi^*(\cdot\vert s)}[\nabla_\theta\log\pi_{\theta_k}(a\vert s)\cdot(\omega_k-\omega^*_k)]-\frac{B\alpha^2}{2}\Vert\omega_k\Vert^2\\		&\overset{(b)}=\alpha[J^{*}-J(\theta_k)]+\alpha\mathbf{E}_{s\sim d^{\pi^*}}\mathbf{E}_{a\sim\pi^*(\cdot\vert s)}\bigg[\nabla_\theta\log\pi_{\theta_k}(a\vert s)\cdot\omega^*_k-A^{\pi_{\theta_k}}(s,a)\bigg]\\
            &+\alpha\mathbf{E}_{s\sim d^{\pi^*}}\mathbf{E}_{a\sim\pi^*(\cdot\vert s)}[\nabla_\theta\log\pi_{\theta_k}(a\vert s)\cdot(\omega_k-\omega^*_k)]-\frac{B\alpha^2}{2}\Vert\omega_k\Vert^2\\
            &\overset{(c)}\geq\alpha[J^{*}-J(\theta_k)]-\alpha\sqrt{\mathbf{E}_{s\sim d^{\pi^*}}\mathbf{E}_{a\sim\pi^*(\cdot\vert s)}\bigg[\bigg(\nabla_\theta\log\pi_{\theta_k}(a\vert s)\cdot\omega^*_k-A^{\pi_{\theta_k}}(s,a)\bigg)^2\bigg]}\\
            &-\alpha\mathbf{E}_{s\sim d^{\pi^*}}\mathbf{E}_{a\sim\pi^*(\cdot\vert s)}\Vert\nabla_\theta\log\pi_{\theta_k}(a\vert s)\Vert_2\Vert(\omega_k-\omega^*_k)\Vert-\frac{B\alpha^2}{2}\Vert\omega_k\Vert^2\\
            &\overset{(d)}\geq\alpha[J^{*}-J(\theta_k)]-\alpha\sqrt{\epsilon_{\mathrm{bias}}}-\alpha G\Vert(\omega_k-\omega^*_k)\Vert-\frac{B\alpha^2}{2}\Vert\omega_k\Vert^2\\
	\end{aligned}	
    \end{equation}
    where the step (a) holds by Assumption \ref{ass_score} and step (b) holds by Lemma \ref{lem_performance_diff}. Step (c) uses the convexity of the function $f(x)=x^2$. Finally, step (d) comes from the Assumption \ref{ass_transfer_error}. Rearranging items, we have
    \begin{equation}
	\begin{split}
            J^{*}-J(\theta_k)&\leq \sqrt{\epsilon_{\mathrm{bias}}}+ G\Vert(\omega_k-\omega^*_k)\Vert+\frac{B\alpha}{2}\Vert\omega_k\Vert^2\\
            &+\frac{1}{\alpha}\mathbf{E}_{s\sim d^{\pi^*}}[KL(\pi^*(\cdot\vert s)\Vert\pi_{\theta_k}(\cdot\vert s))-KL(\pi^*(\cdot\vert s)\Vert\pi_{\theta_{k+1}}(\cdot\vert s))]
	\end{split}
    \end{equation}
    Summing from $k=1$ to $K$, using the non-negativity of KL divergence and dividing the resulting expression by $K$, we get the desired result.
\end{proof}
	

\subsection{Proof of Lemma \ref{lemma:41ss}}

\begin{proof}
    By the $L$-smooth property of the objective function, we have
    \begin{align}
        \begin{split}
            J(\theta_{k+1})&\geq J(\theta_k)+\left<\nabla J(\theta_k),\theta_{k+1}-\theta_k\right>-\frac{L}{2}\Vert\theta_{k+1}-\theta_k\Vert^2\\
            &\overset{(a)} =J(\theta_k) + \alpha \nabla J(\theta_k)^T \omega_k - \frac{L \alpha^2}{2} \norm{ \omega_k }^2 \\
            &= J(\theta_k) + \alpha \norm{\nabla J(\theta_k) }^2 - \alpha \langle \nabla J(\theta_k) - \omega_k, \nabla J(\theta_k) \rangle - \frac{L \alpha^2}{2}\Vert \nabla J(\theta_k)-\omega_k-\nabla J(\theta_k)\Vert^2\\
            &\overset{(b)}\geq J(\theta_k) + \alpha \norm{\nabla J(\theta_k) }^2 - \frac{\alpha}{2} \Vert\nabla J(\theta_k) - \omega_k\Vert^2 -\frac{\alpha}{2}\Vert\nabla J(\theta_k)\Vert^2\\
            &\quad- L\alpha^2\Vert \nabla J(\theta_k)-\omega_k\Vert^2-L\alpha^2\Vert\nabla J(\theta_k)\Vert^2\\
            &= J(\theta_k) + \left(\frac{\alpha}{2}-L\alpha^2\right) \norm{\nabla J(\theta_k) }^2 - \left(\frac{\alpha}{2}+L\alpha^2\right) \Vert\nabla J(\theta_k) - \omega_k\Vert^2
        \end{split}
    \end{align}
    where step (a) holds from the fact that $\theta_{t+1} = \theta_k + \alpha \omega_k$ and step (b) holds due to the Cauchy-Schwarz inequality. Rearranging the terms yields the following. 

    \begin{equation}
        \norm{ \nabla J(\theta_k) }^2 \leq \frac{J(\theta_{k+1}) - J(\theta_k) + (\frac{\alpha}{2}+L\alpha^2) \Vert\nabla J(\theta_k) - \omega_k\Vert^2}{\frac{\alpha}{2}-L\alpha^2} 
    \end{equation}

    Choosing $\alpha = \frac{1}{4L}$ and summing over $k\in\{ 1, 2, \cdots, K\}$ results in the following.

    \begin{align}
        \dfrac{1}{K}\sum_{k=1}^{K} \norm{ \nabla J(\theta_k) }^2 \leq \dfrac{16L}{K}\left[J(\theta_K)-J(\theta_0)\right]+  \dfrac{3}{K}\sum_{k=1}^{K}\Vert\nabla J(\theta_k) - \omega_k\Vert^2 
    \end{align}

    Using $|J(\theta_k)|\leq 1$ due to bounded reward, we conclude the result.
\end{proof}


\section{Proofs for the Section of Regret Analysis}


\subsection{Proof of Lemma \ref{lemma_last}}

\begin{proof}
    Using Taylor's expansion, we can write the following $\forall (s, a)\in \mathcal{S}\times \mathcal{A}$, $\forall k$.
    \begin{align}
        \label{eq_pi_lipschitz}
        \begin{split}
            |\pi_{\theta_{k+1}}(a|s)-\pi_{\theta_{k}}(a|s)|&=\left|(\theta_{k+1}-\theta_k)^T\nabla_{\theta}\pi_{\bar\theta}(a|s) \right| \\&=\pi_{\bar{\theta}_k}(a|s)\left|(\theta_{k+1}-\theta_k)^T\nabla_{\theta}\log \pi_{\bar{\theta}_k}(a|s) \right|\\
            &\leq \pi_{\bar{\theta}_k}(a|s) \norm{\theta_{k+1}-\theta_k}\norm{\nabla_{\theta}\log \pi_{\bar{\theta}_k}(a|s)}\overset{(a)}{\leq} G\pi_{\bar{\theta}_k}(a|s)\norm{\theta_{k+1}-\theta_k}
        \end{split} 
    \end{align}
    where $\bar{\theta}_k$ is some convex combination of $\theta_{k}$ and $\theta_{k+1}$ and $(a)$ follows from Assumption \ref{ass_score}. This concludes the first statement. Applying \eqref{eq_pi_lipschitz} and Lemma \ref{lemma_aux_5}, we obtain,

    \begin{align}
        \label{eq_long_49}
        \begin{split}
            \sum_{k=1}^{K}\mathbf{E}\left|J(\theta_{k+1}) - J(\theta_{k})\right| &= \sum_{k=1}^{K}\mathbf{E}\left|\sum_{s,a}d^{\pi_{\theta_{k+1}}}(s)(\pi_{\theta_{k+1}}(a|s)-\pi_{\theta_{k}}(a|s))Q^{\pi_{\theta_{k}}}(s, a)\right|\\
            &\leq \sum_{k=1}^{K}\mathbf{E}\left[\sum_{s,a}d^{\pi_{\theta_{k+1}}}(s)\left|\pi_{\theta_{k+1}}(a|s)-\pi_{\theta_{k}}(a|s)\right|\left|Q^{\pi_{\theta_{k}}}(s, a)\right|\right]\\
            &\leq G\sum_{k=1}^{K}\mathbf{E}\left[\sum_{s,a}d^{\pi_{\theta_{k+1}}}(s)\pi_{\bar{\theta}_k}(a|s)\Vert\theta_{k+1}-\theta_{k}\Vert|Q^{\pi_{\theta_{k-1}}}(s, a)|\right]\\
            &\overset{(a)}{\leq} G\alpha\sum_{k=1}^{K}\mathbf{E}\left[\underbrace{\sum_{s,a}d^{\pi_{\theta_{k+1}}}(s)\pi_{\bar{\theta}_k}(a|s)}_{=1}\Vert\omega_k\Vert\cdot 6t_{\mathrm{mix}}\right]\\
            &\overset{}{=} 6G\alpha t_{\mathrm{mix}} \sum_{k=1}^{K}\mathbf{E}\norm{\omega_k}\\
            &\overset{(b)}\leq 6G\alpha t_{\mathrm{mix}}\sqrt{K}\left(\sum_{k=1}^{K}\mathbf{E}\norm{\omega_k}^2\right)^{\frac{1}{2}}\overset{(c)}{\leq} \mathcal{\Tilde{O}}\left(\dfrac{\alpha G T^{\frac{1}{4}}}{t_{\mathrm{hit}}}\left[\sqrt{A}G t_{\mathrm{mix}}+\sqrt{L t_{\mathrm{mix}}t_{\mathrm{hit}}}\right]\right)
        \end{split}
    \end{align}
 
    Inequality $(a)$ uses Lemma \ref{lemma_aux_2} and the update rule $\theta_{k+1}=\theta_k+\alpha \omega _k$. Step $(b)$ holds by the Cauchy inequality and Jensen inequality whereas $(c)$ can be derived using $\eqref{eq_34}$ and substituting $K=T/H$. This establishes the second statement. Next, recall from $(\ref{eq_r_pi_theta})$ that for any policy $\pi_{\theta}$,  $r^{\pi_{\theta}}(s) \triangleq \sum_a\pi_{\theta}(a|s)r(s, a)$.
    Note that, for any policy parameter $\theta$, and any state $s\in\mathcal{S}$, the following holds.
    \begin{align}
        \label{eq_49}
        V^{\pi_{\theta}}(s)=\sum_{t=0}^{\infty}\left<(P^{\pi_{\theta}})^t(s,\cdot) - d^{\pi_{\theta}}, r^{\pi_{\theta}}\right> = \sum_{t=0}^{N-1}\left<(P^{\pi_{\theta}})^t({s,\cdot}),r^{\pi_{\theta}}\right> - NJ(\theta) + \sum_{t=N}^{\infty}\left<(P^{\pi_{\theta}})^t({s,\cdot}) - d^{\pi_{\theta}},r^{\pi_{\theta}}\right>.
    \end{align}

    Define the following quantity.
    \begin{align}
        \label{def_error}
        \delta^{\pi_{\theta}}(s, T) \triangleq \sum_{t=N}^{\infty}\norm{(P^{\pi_{\theta}})^t({s,\cdot}) - d^{\pi_{\theta}}}_1 ~~\text{where} ~N=4t_{\mathrm{mix}}(\log_2 T)
    \end{align}
 
    Lemma \ref{lemma_aux_3} states that for sufficiently large $T$, we have $\delta^{\pi_{\theta}}(s, T)\leq \frac{1}{T^3}$ for any policy $\pi_{\theta}$ and state $s$. Combining this result with the fact that the reward function is bounded in $[0, 1]$, we obtain,
    \begin{align}
        \label{eq_exp_diff_v}
	\begin{split}    
            &\sum_{k=1}^K\mathbf{E}|V^{\pi_{\theta_{k+1}}}(s_k) - V^{\pi_{\theta_{k}}}(s_k)|\\
            &\leq \sum_{k=1}^K\mathbf{E}\left|\sum_{t=0}^{N-1}\left<(P^{\pi_{\theta_{k+1}}})^t({s_k,\cdot}) - (P^{\pi_{\theta_k}})^t({s_k,\cdot}), r^{\pi_{\theta_{k+1}}}\right>\right| + \sum_{k=1}^K\mathbf{E}\left|\sum_{t=0}^{N-1}\left<(P^{\pi_{\theta_k}})^t({s_k,\cdot}), r^{\pi_{\theta_{k+1}}}-r^{\pi_{\theta_k}}\right>\right| \\
            & \hspace{8cm}+ N\sum_{k=1}^K\mathbf{E}|J(\theta_{k+1}) - J(\theta_{k})| + \frac{2K}{T^3}\\
            &\overset{(a)}{\leq} \sum_{k=1}^K\sum_{t=0}^{N-1}\mathbf{E}\norm{ (P^{\pi_{\theta_{k+1}}})^t - (P^{\pi_{\theta_k}})^t)r^{\pi_{\theta_{k+1}}} }_{\infty} + \sum_{k=1}^K\sum_{t=0}^{N-1}\mathbf{E}\norm{r^{\pi_{\theta_{k+1}}}-r^{\pi_{\theta_k}}}_{\infty} \\
            & \hspace{7cm}+ \mathcal{\Tilde{O}}\left(\dfrac{\alpha G T^{\frac{1}{4}}t_{\mathrm{mix}}}{t_{\mathrm{hit}}}\left[\sqrt{A}G t_{\mathrm{mix}}+\sqrt{L t_{\mathrm{mix}}t_{\mathrm{hit}}}\right]\right)
        \end{split}
    \end{align}
    where $(a)$ follows from \eqref{eq_long_49} and substituting $N=4t_{\mathrm{mix}}(\log_2 T)$. For the first term, note that,
    \begin{align}
        \label{eq_long_recursion}
        \begin{split}
            &\norm{ ((P^{\pi_{\theta_{k+1}}})^t - (P^{\pi_{\theta_k}})^t)r^{\pi_{\theta_{k+1}}} }_{\infty}\\ &\leq \norm{ P^{\pi_{\theta_{k+1}}}((P^{\pi_{\theta_{k+1}}})^{t-1} - (P^{\pi_{\theta_k}})^{t-1})r^{\pi_{\theta_{k+1}}} }_{\infty} + \norm{ (P^{\pi_{\theta_{k+1}}} - P^{\pi_{\theta_k}})(P^{\pi_{\theta_k}})^{t-1}r^{\pi_{\theta_{k+1}}} }_{\infty}\\
            &\overset{(a)}{\leq} \norm{ ((P^{\pi_{\theta_{k+1}}})^{t-1} - (P^{\pi_{\theta_k}})^{t-1})r^{\pi_{\theta_{k+1}}} }_{\infty} + \max_s\norm{P^{\pi_{\theta_{k+1}}}({s,\cdot})-P^{\pi_{\theta_k}}({s,\cdot})}_1
        \end{split}
    \end{align}

    Inequality $(a)$ holds since every row of $P^{\pi_{\theta_k}}$ sums to $1$ and $\norm{(P^{\pi_{\theta_k}})^{t-1}r^{\pi_{\theta_{k+1}}}}_{\infty}\leq 1$. Moreover, invoking \eqref{eq_pi_lipschitz}, and the parameter update rule $\theta_{k+1}=\theta_k + \alpha \omega_k$, we get,
    \begin{align*}
        \max_s\Vert P^{\pi_{\theta_{k+1}}}({s,\cdot})-P^{\pi_{\theta_k}}({s,\cdot})\Vert_1 &= \max_s\left| \sum_{s'}\sum_a(\pi_{\theta_{k+1}}(a|s)-\pi_{\theta_k}(a|s))P(s'|s, a)\right| \\
        &\leq G \norm{\theta_{k+1}-\theta_k}\max_s\left| \sum_{s'}\sum_a \pi_{\bar{\theta}_k}(a|s)P(s'|s, a)\right| \\
        &\leq \alpha G\norm{\omega_k}  
    \end{align*}
 
    Plugging the above result into $(\ref{eq_long_recursion})$ and using a recursive argument, we get,
    \begin{align*}
        \norm{ ((P^{\pi_{\theta_{k+1}}})^t - (P^{\pi_{\theta_k}})^t)r^{\pi_{\theta_{k+1}}} }_{\infty} &\leq \sum_{t'=1}^{t} \max_s\norm{P^{\pi_{\theta_{k+1}}}({s,\cdot})-P^{\pi_{\theta_k}}({s,\cdot})}_1\\
        &\leq \sum_{t'=1}^{t}\alpha G\norm{\omega_k}  \leq \alpha t G\norm{\omega_k}
    \end{align*}
    
    Finally, we have
    \begin{align}
        \label{eq_app_54}
        \begin{split}
            \sum_{k=1}^K\sum_{t=0}^{N-1} &\mathbf{E}\norm{ ((P^{\pi_{\theta_{k+1}}})^t - (P^{\pi_{\theta_k}})^t)r^{\pi_{\theta_{k+1}}} }_{\infty}\\
            &\leq \sum_{k=1}^K\sum_{t=0}^{N-1}\alpha t G\norm{\omega_k}\\
            &\leq \mathcal{O}(\alpha G N^2) \sum_{k=1}^K\mathbf{E}\norm{\omega_k}\\
            &\leq \mathcal{O}(\alpha G N^2\sqrt{K}) \left(\sum_{k=1}^K\mathbf{E}\norm{\omega_k}^2\right)^{\frac{1}{2}}\\
            &\overset{(a)}{\leq} \mathcal{\Tilde{O}}\left(\dfrac{\alpha G T^{\frac{1}{4}}t_{\mathrm{mix}}}{t_{\mathrm{hit}}}\left[\sqrt{A}G t_{\mathrm{mix}}+\sqrt{L t_{\mathrm{mix}}t_{\mathrm{hit}}}\right]\right)
        \end{split}
    \end{align}
    where $(a)$ follows from \eqref{eq_34}. Moreover, notice that,
    \begin{equation}
        \label{eq_app_55}
        \begin{aligned}
            \sum_{k=1}^{K}\sum_{t=0}^{N-1}\mathbf{E}\norm{r^{\pi_{\theta_{k+1}}}-r^{\pi_{\theta_{k}}}}_{\infty}&\overset{}{\leq} \sum_{k=1}^{K}\sum_{t=0}^{N-1}\mathbf{E}\left[\max_s\left|\sum_a(\pi_{\theta_{k+1}}(a|s)-\pi_{\theta_{k}}(a|s))r(s,a)\right|\right]\\
            &\overset{(a)}{\leq}\alpha GN \sum_{k=1}^{K} \mathbf{E}\norm{\omega_k}\\
            &\leq \alpha GN\sqrt{K} \left(\sum_{k=1}^{K} \mathbf{E}\norm{\omega_k}^2\right)^{\frac{1}{2}}\\
            &\overset{(b)}{\leq} \mathcal{\Tilde{O}}\left(\dfrac{\alpha G T^{\frac{1}{4}}}{t_{\mathrm{hit}}}\left[\sqrt{A}G t_{\mathrm{mix}}+\sqrt{L t_{\mathrm{mix}}t_{\mathrm{hit}}}\right]\right)
        \end{aligned}
    \end{equation}
    where $(a)$ follows from \eqref{eq_pi_lipschitz} and the parameter update rule $\theta_{k+1}=\theta_k + \alpha \omega_k$ while $(b)$ is a consequence of \eqref{eq_34}. Combining \eqref{eq_exp_diff_v}, \eqref{eq_app_54}, and  \eqref{eq_app_55}, we establish the third statement.
\end{proof}


\subsection{Proof of Theorem \ref{thm_regret}}

Recall the decomposition of the regret in Eq. \eqref{reg_decompose} and take the expectation
\begin{align}
    \begin{split}
        \mathbf{E}[\mathrm{Reg}_T] &= \sum_{t=0}^{T-1} \left(J^* - r(s_t, a_t)\right)=H\sum_{k=1}^{K}\left(J^*-J({\theta_k})\right)+\sum_{k=1}^{K}\sum_{t\in\mathcal{I}_k} \left(J(\theta_k)-r(s_t, a_t)\right)\\
        &=H\sum_{k=1}^{K}\left(J^*-J({\theta_k})\right)+\mathbf{E}\left[\sum_{k=1}^{K-1} V^{\pi_{\theta_{k+1}}}(s_{kH})-V^{\pi_{\theta_k}}(s_{kH})\right]+\mathbf{E}\left[ V^{\pi_{\theta_K}}(s_{T})-V^{\pi_{\theta_0}}(s_{0})\right]
    \end{split}
\end{align}
and using the result in  (\ref{thm_final_convergence}), Lemma \ref{lemma_last} and Lemma \ref{lemma_aux_2}, we get,
\begin{align}
    \begin{split}
        \mathbf{E}&[\mathrm{Reg}_T]\leq \mathcal{\tilde{O}}\left(\dfrac{ABG^2t_{\mathrm{mix}}^2+BLt_{\mathrm{mix}}t_{\mathrm{hit}}}{L}\sqrt{T}\right) + \mathcal{\tilde{O}}\left(\left[\sqrt{A}G^2t_{\mathrm{mix}}+G\sqrt{Lt_{\mathrm{mix}}t_{\mathrm{hit}}}\right]\left(1+\dfrac{1}{\mu_F}\right)T^{\frac{3}{4}}\right)+  T\sqrt{\epsilon_{\mathrm{bias}}}\\
        &+\mathcal{\tilde{O}}\left(Lt_{\mathrm{mix}}t_{\mathrm{hit}}\mathbf{E}_{s\sim d^{\pi^*}}[KL(\pi^*(\cdot\vert s)\Vert\pi_{\theta_1}(\cdot\vert s))]\sqrt{T}\right)+\mathcal{\tilde{O}}\left(\dfrac{ t_{\mathrm{mix}}}{Lt_{\mathrm{hit}}}\left[\sqrt{A}G^2t_{\mathrm{mix}}+G\sqrt{Lt_{\mathrm{mix}}t_{\mathrm{hit}}}\right]T^{\frac{1}{4}}\right)+\mathcal{O}(t_{\mathrm{mix}})
    \end{split}
\end{align}


\section{Some Auxiliary Lemmas for the Proofs}

\begin{lemma}
    \label{lemma_aux_2}
    \citep[Lemma 14]{wei2020model} For any ergodic MDP with mixing time $t_{\mathrm{mix}}$, the following holds $\forall (s, a)\in\mathcal{S}\times \mathcal{A}$ and any policy $\pi$.
    \begin{align*}
        (a) |V^{\pi}(s)|\leq 5 t_{\mathrm{mix}},~~
        (b) |Q^{\pi}(s, a)|\leq 6 t_{\mathrm{mix}}
    \end{align*}
\end{lemma}

\begin{lemma}
    \label{lemma_aux_3}
    \citep[Corollary 13.2]{wei2020model} Let $\delta^{\pi}(\cdot, T)$ be defined as written below for an arbitrary policy $\pi$. 
    \begin{align}
        \label{def_error_aux_1}
        \delta^{\pi}(s, T) \triangleq \sum_{t=N}^{\infty}\norm{(P^{\pi})^t({s,\cdot}) - d^{\pi}}_1, ~\forall s\in\mathcal{S} ~~\text{where} ~N=4t_{\mathrm{mix}}(\log_2 T)
    \end{align}
    
    If $t_{\mathrm{mix}}<T/4$, we have the following inequality $\forall s\in\mathcal{S}$: $\delta^{\pi}(s, T)\leq \frac{1}{T^3}$.
\end{lemma}

\begin{lemma}
    \label{lemma_aux_4}
    \citep[Lemma 16]{wei2020model} Let $\mathcal{I}=\{t_1+1,t_1+2,\cdots,t_2\}$ be a certain period of an epoch $k$ of Algorithm \ref{alg:estQ} with length $N$. Then for any $s$, the probability that the algorithm never visits $s$ in $\mathcal{I}$ is upper bounded by
    \begin{equation}
        \left(1-\frac{3d^{\pi_{\theta_k}}(s)}{4}\right)^{\left\lfloor\frac{\lfloor \mathcal{I}\rfloor}{N}\right\rfloor}
    \end{equation}
\end{lemma}

\begin{lemma}
    \label{lemma_aux_5}
    \citep[Lemma 15]{wei2020model} For two difference policy $\pi$ and $\pi'$, the difference of the objective function $J$ is
    \begin{equation}
        J^{\pi}-J^{\pi'}=\sum_{s}\sum_{a}d^{\pi}(s)(\pi(a|s)-\pi'(a|s))Q^{\pi'}(s,a)
    \end{equation}
\end{lemma}

\begin{lemma}
    \label{lemma_aux_6}
    \citep[Lemma A.6]{dorfman2022adapting} Let  $\theta\in\Theta$ be a policy parameter. Fix a trajectory $z=\{(s_t, a_t, r_t, s_{t+1})\}_{t\in\mathbb{N}}$ generated by following the policy $\pi_{\theta}$ starting from some initial state $s_0\sim\rho$. Let, $\nabla L(\theta)$ be the gradient that we wish to estimate over $z$, and $l(\theta, \cdot)$ is a function such that $\mathbf{E}_{z\sim d^{\pi_{\theta}}, \pi_{\theta}}l(\theta, z)=\nabla L(\theta)$. Assume that $\norm{l(\theta, z)}, \norm{\nabla L(\theta)}\leq G_L$, $\forall \theta\in\Theta$, $\forall z\in \mathcal{S}\times \mathcal{A}\times \mathbb{R}\times \mathcal{S}$. Define $l^{Q}=\frac{1}{Q}\sum_{i=1}^Q l(\theta, z_i)$. If $P=2t_{\mathrm{mix}}\log T$, then the following holds as long as $Q\leq T$,
    \begin{align}
        \mathbf{E}\left[\norm{l^{Q}-\nabla L(\theta)}^2\right]\leq \mathcal{O}\left(G_L^2\log\left(PQ\right)\dfrac{P}{Q}\right)
    \end{align}
\end{lemma}